\pdfoutput=1

\documentclass[11pt]{article}

\usepackage[review]{ACL2023}

\usepackage{times}
\usepackage{latexsym}
\usepackage{tikz}
\usepackage{hyperref}
\usepackage[edges]{forest}
\usepackage{graphicx}
\usepackage{algorithm}
\usepackage[noend]{algpseudocode}
\algnewcommand{\LeftComment}[1]{\Statex \(\triangleright\) #1}
\usepackage{multicol}
\usepackage{rotating}
\usepackage{tikz}
\usetikzlibrary{trees,shapes,calc}
\usepackage{makecell}
\let\OldMakecell\makecell
\renewcommand{\makecell}[1]{%
    \OldMakecell{\color{.}{#1}} 
}
\usepackage[T1]{fontenc}

\usepackage[utf8]{inputenc}

\usepackage{microtype}
\usepackage{amsmath}
\usepackage{inconsolata}

\usepackage{titlesec}
\titlespacing{\section}{0pt}{-1pt}{-1pt} 
\titlespacing{\subsection}{0pt}{-1pt}{-1pt} 
\titlespacing\subsubsection{0pt}{0pt}{0pt}
\usepackage{mdframed}
\usepackage[textsize=small]{todonotes}

\usepackage{booktabs}
\usepackage{rotating}

\usepackage{tcolorbox}
\usepackage{colortbl}
\usepackage{adjustbox}
\colorlet{lightgray}{White!70!lightgray}
\colorlet{lightblue}{White!10!MidnightBlue}

\newcommand{\ignore}[1]{}

\title{A Systematic Survey of Automatic Prompt Optimization Techniques}

\author{Kiran Ramnath, Kang Zhou, {\bf Sheng Guan}, {\bf Soumya Smruti Mishra}, {\bf Xuan Qi}, {\bf Zhengyuan Shen}, 
\\ {\bf Shuai Wang}, {\bf Sangmin Woo}, {\bf Sullam Jeoung}, {\bf Yawei Wang}, {\bf Haozhu Wang}, {\bf Han Ding},   
\\ {\bf Yuzhe Lu}, {\bf Zhichao Xu}, {\bf Yun Zhou}, {\bf Balasubramaniam Srinivasan}, {\bf Qiaojing Yan}, {\bf Yueyan Chen}, 
\\ {\bf Haibo Ding}, {\bf Panpan Xu}, \and {\bf Lin Lee Cheong}\\
  Amazon Web Services \\
  \parbox[c]{\textwidth}{\texttt{\{raxkiran,zhoukang,shguan,soumish,xuaqi,donshen, wshui,sangminw,sullamij, \\
  yawenwan, haozhuw, handing, yuzhelu, xzhichao, yunzzhou, srbalasu, qiaojiny, yyanc, hbding, xupanpan, lcheong\}@amazon.com}}}
\begin{document}
\maketitle
\begin{abstract}
Since the advent of large language models (LLMs), prompt engineering has been a crucial step for eliciting desired responses for various Natural Language Processing (NLP) tasks. However, prompt engineering remains an impediment for end users due to rapid advances in models, tasks, and associated best practices. To mitigate this, Automatic Prompt Optimization (APO) techniques have recently emerged that use various automated techniques to help improve the performance of LLMs on various tasks. In this paper, we present a comprehensive survey summarizing the current progress and remaining challenges in this field. We provide a formal definition of APO, a 5-part unifying framework, and then proceed to rigorously categorize all relevant works based on their salient features therein. We hope to spur further research guided by our framework. 
\end{abstract}
\section{Introduction}
\label{sec:introduction}
Since \citet{decanlp} cast multi-task NLP as Question Answering, using prompts as inputs has become the standard way to elicit desired responses from Large Language Models (LLMs). Furthermore, LLMs' few-shot learning \cite{fewshot}, instruction-following \cite{instruction-alignment}, and zero-shot reasoning capabilities \cite{zeroshot} have led to a widespread proliferation of prompting tricks for various tasks and model variants. However, LLMs still exhibit unpredictable sensitivity to various factors (explanation of the task \cite{li2023large},ordering \cite{liu2024lost}, stylistic formatting \cite{sclarquantifying}, etc.) causing a performance gap between two prompts that are semantically similar, thereby adding impediments for adoption by end users. Against this backdrop, Black-Box Automatic Prompt Optimization (APO) techniques have emerged that improve task performance via automated prompt improvements. The possess various attractive features - (1) they do not require parameter access on LLMs performing the task, (2) they systematically search through the prompt solution space, and (3) they retain human interpretability of prompt improvements. \ignore{\citet{smart} categorizes two broad directions in APO - Exemplar Optimization (EO, also called In-context learning) and Instruction Optimization (IO).\ignore{\todo{This sounds like the survey paper only focuses on IO? Will that be systematic then as the title is named?}} While \citet{icl-survey} surveyed EO in detail, our survey fills this gap for IO.} In this survey paper, we aim to highlight the advances in the field. Our core contribution is a 5-part APO taxonomy combined with a comprehensive fine-grained categorization of various design choices therein (see Fig. \ref{main-figure}, Tables \ref{tab:comparison_1}, \ref{tab:comparison_2}, \ref{tab:comparison_3} in Appendix). We hope our framework will be informational for new and seasoned researchers alike, enabling further research on open questions. 



\begin{figure*}[ht] 
\resizebox*{\textwidth}{.66\textheight}
{%
\begin{forest}
for tree={
    grow=east,
    font=\large,
    parent anchor=east,
    anchor=west,
    edge={->, rounded corners},
    edge path={
        \noexpand\path [draw, \forestoption{edge}] (!u.parent anchor) -- ++(3mm,0) |- (.child anchor)\forestoption{edge label};
    },
    inner sep=1mm,
    text width=4cm,
    l sep=10mm,
    text centered,
    align=center,
    rounded corners, 
    draw=black 
},
for children={%
    tier/.wrap pgfmath arg={level#1}{level()}, 
    s sep+=5mm, 
}
[Prompt optimization anatomy \textsection \ref{anatomy}, rectangle,fill=violet!20,rotate=90,parent anchor=south,text width=7cm, align=c,
    [Iteration depth \textsection \ref{depth}, rectangle, fill=cyan!20, text width=6cm,
        [Variable steps \textsection\ref{variable}, text width=15.22cm,fill=cyan!20],
        [Fixed steps \textsection\ref{fixed}, text width=15.22cm,fill=cyan!20]
    ],
    [Filter and retain \\promising candidates \textsection \ref{filter}, rectangle, fill=pink!25,  text width=6cm,
        [Meta-heuristic ensemble \textsection\ref{metaheuristic}, text width=15.22cm, fill=pink!25],
        [Region-based joint search \textsection\ref{rbjs}, text width=15.22cm, fill=pink!25],
        [Upper confidence bound and variants \textsection\ref{ucb}, text width=15.22cm, fill=pink!25],
        [TopK Greedy Search \textsection\ref{topk}, text width=15.22cm, fill=pink!25],
    ],
    [Candidate prompt \\generation \textsection \ref{candidate}, rectangle, fill=blue!10,  text width=6cm, 
        [{Program Synthesis \textsection \ref{program}}, text width = 15.22cm,fill=blue!10],
        [Coverage-based \textsection \ref{coverage},fill=blue!10,
            [Ensemble methods \textsection \ref{ensemble}, text width=10cm,fill=blue!10],
            [Mixture of experts \textsection \ref{moe}, text width=10cm,fill=blue!10],
            [Single prompt expansion \textsection \ref{expand}, text width=10cm,fill=blue!10]
        ],
        [Metaprompt design \textsection \ref{metaprompt}, text width = 15.22cm,fill=blue!10],
        [Editing with auxiliary \\trained NN \textsection \ref{nn},fill=blue!10,
            [Generative Adversarial Networks \textsection \ref{gan}, text width=10cm,fill=blue!10],
            [LLM Finetuning \textsection \ref{ft}, text width=10cm,fill=blue!10],
            [Reinforcement Learning \textsection \ref{rl}, text width=10cm,fill=blue!10]
        ],        
        [Heuristic-based \\edits \textsection \ref{heuristic},fill=blue!10,
            [Vocabulary pruning \textsection \ref{prune},fill=blue!10,text width=10cm],   
            [Word / phrase edits \textsection \ref{word}, fill=blue!10, text width=10cm],
            [Genetic Algorithm \textsection \ref{genetic},fill=blue!10, text width=10cm 
            ],
            [Monte Carlo Sampling \textsection \ref{mc},fill=blue!10,text width=10 cm]            
        ]
    ], 
    [Inference evaluation \\and feedback \textsection \ref{evaluate}, rectangle, fill=yellow!50, text width=6cm,
        [Human Feedback \textsection \ref{human}, fill=yellow!50, text width=15.22cm],
        [LLM Feedback \textsection \ref{llm-feedback},rectangle, fill=yellow!50,
            [Improving multiple candidates \textsection \ref{multiple}, text width=10cm,fill=yellow!50],
            [Improving single candidate \textsection \ref{single}, text width=10cm,fill=yellow!50]
        ],
        [Numeric score \textsection \ref{score-feedback},rectangle, fill=yellow!50,
            [Negative log-likelihood \textsection\ref{nll}, text width=10cm,fill=yellow!50],
            [Entropy-based \textsection\ref{entropy}, text width=10cm,fill=yellow!50],
            [Reward model score \textsection\ref{reward}, text width=10cm,fill=yellow!50],
            [Task accuracy \textsection\ref{accuracy}, text width=10cm,fill=yellow!50],
        ]
    ],
    [Seed Prompts \textsection \ref{seed}, rectangle, fill=orange!30, text width=6cm,
        [Instruction-induction via LLMs \textsection\ref{induction}, fill=orange!30, text width=15.22cm],
        [Manual Instructions \textsection\ref{manual}, fill=orange!30,text width=15.22cm]
    ]
]
\end{forest}
}
\caption{Taxonomy of Automatic Prompt Optimization} 
\label{main-figure}
\end{figure*}
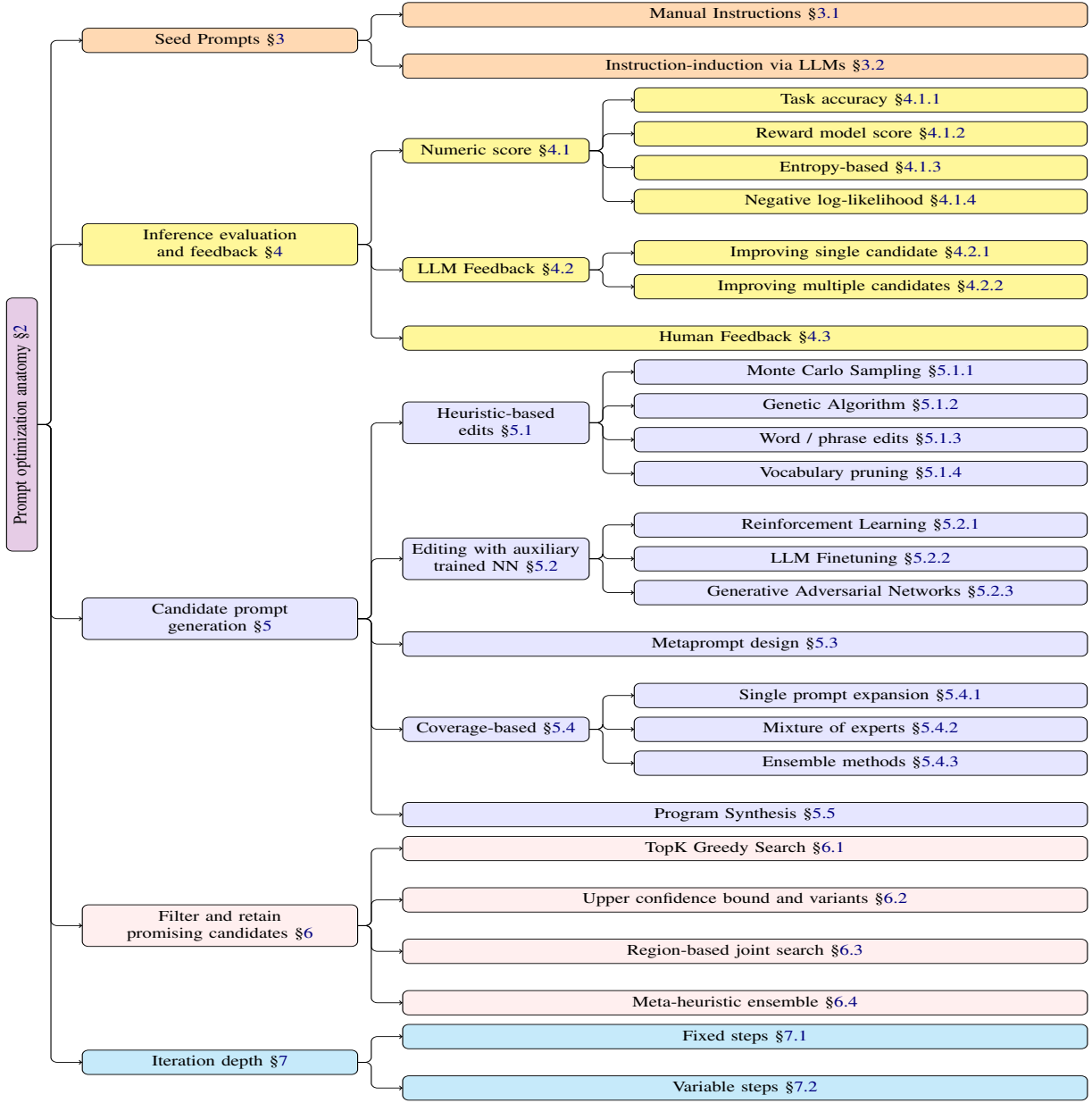

\section{Automatic Prompt Optimization Formulation}
\label{anatomy}
We formalize the process of automatic prompt optimization (APO) as follows. Given a task model $M_{task}$, initial prompt $\rho$ $\in V$, the goal of an APO-system $M_{APO}$ is to obtain the best performing prompt-template $\rho^{opt}$ under a metric $f \in F$ and eval-set $D_{val}$ 
\begin{align}{
\rho^{opt}:= \arg\max_{\substack{\rho \in V}}E_{x \sim D_{val}}[f(M_{task} (\rho \oplus x))]
}
\end{align}
This objective function is not tractable for discrete prompt optimization as token-sequence search spaces are combinatorial. Instead, APO techniques follow the general anatomy as described in Algorithm \ref{anatomy-algo} to obtain approximate solutions. 

\begin{algorithm}
\caption{Prompt optimization framework} 
\label{anatomy-algo}
\begin{algorithmic}[1]
\State $P_0 := \{\rho_1, \rho_2, \ldots, \rho_k\}$ \Comment{\textsection\ref{seed}. \text{\color{gray}Seed prompts}}
\State $D_{val} := \{(x_1, y_1)\}_{i=1}^n$ \Comment{\text{\color{gray}Validation set}}
\State $f_1,\ldots,f_m \in F$ \Comment{\textsection\ref{evaluate}. \text{\color{gray}Inference evaluation}}
\For {$t=1,2,\ldots,N$} \Comment{\textsection\ref{depth}. \text{\color{gray}Iteration depth}}
\LeftComment{\textsection\ref{candidate}.  \text{\color{gray}Generate prompt candidates}}
    \State $G_t :=M_{_{APO}}(P, D_{val}, F)$
    \LeftComment{\textsection\ref{filter}.  \text{\color{gray}Filter and retain candidates}}
    \State $P_t := Select(G_t,  D_{val}, F)$          
    \LeftComment{\textsection\ref{depth}.  \text{\color{gray}{Optionally check for early convergence}}}
    \If{$f_{convergence} \leq \epsilon$}
        \State\textbf{exit} 
    \EndIf
\EndFor
\State \Return $\arg\max_{\substack{\rho \in P_N}} E_{\substack{x \sim D_{val}}}\left[f(M_{task}(\rho \oplus x))\right]$
\end{algorithmic} 
\end{algorithm}
\begin{figure*}[htbp]
    \centering
\includegraphics[width=\textwidth, height=4cm]{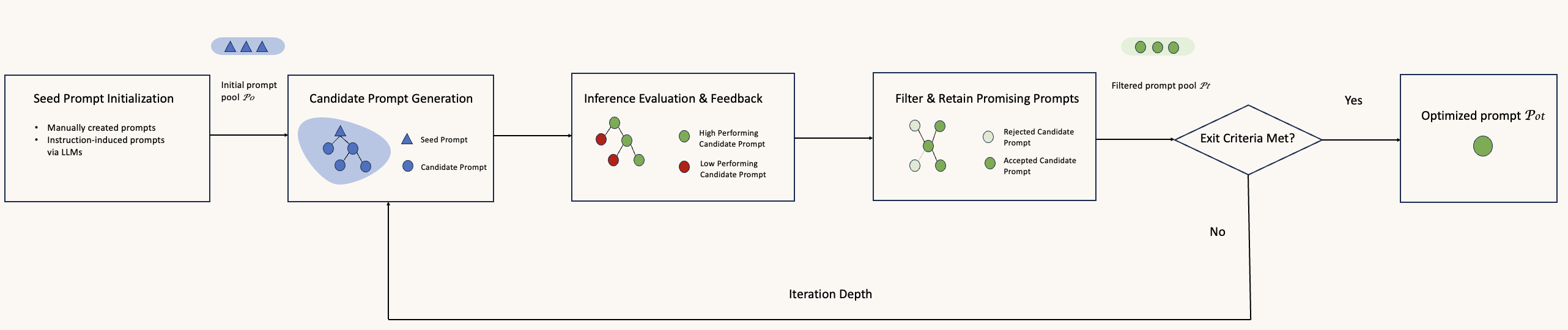}
\caption{Representative APO system}
\label{fig:system-diagram}
\end{figure*}
\section{Initialize Seed Prompts}
\label{seed}

\subsection{Manual Instructions}
\label{manual}
Several approaches use a seed of manually created instructions that offer interpretable and strong baselines as the basis for further improvement,\textit{inter alia.}, ProteGi \cite{protegi}, GPS \cite{gps}, SPRIG \cite{sprig}. While obtaining quality examples can be costly, APE \cite{ape} \footnote{Note: APE stands for Automatic Prompt Engineer method introduced by \cite{ape}, not to be confused with APO which broadly refers to the entire area of Automatic Prompt Optimization} showed that a few hundred samples are sufficient for further optimization. 

\subsection{Instruction Induction via LLMs}
\label{induction}
\citet{instruction-induction} were the first to propose inducing LLMs to infer human-readable prompts based on a few demonstrations $E$ (see Appendix \ref{appendix:instruction_induction} for prompt). APE \cite{ape} and DAPO \cite{dapo} use the induced seed instructions for further optimization, while MOP \cite{mop} and GPO \cite{gpo} use APE to induce cluster-specific prompts. Apart from demonstrations, SCULPT \cite{sculpt} induced instructions from task-READMEs, while UniPrompt \cite{uniprompt} used LLMs to fill-in structured templates. 

\begin{table*}[ht]
\rowcolors{2}{gray!15}{white} 
\resizebox{1\textwidth}{!}{
\begin{tabular}{p{3cm}p{4cm}p{2.5cm}p{3.5cm}p{3.5cm}p{4cm}}
\hline
\textbf{Paper} & \textbf{Seed instructions} & \textbf{Iteration depth} & \textbf{Inference evaluation} & \textbf{Candidate generation} & \textbf{Search+filter strategy}\\
\hline
    ProTeGi \cite{protegi} &Manually created & Fixed & LLM feedback + \newline Task accuracy &LLM rewriter & UCB for trees \\
    APE \cite{ape}  &  Instruction induction & Fixed & Task accuracy & N/A & UCB \\
    CRISPO \cite{crispo} &Manually created &Fixed &LLM feedback +\newline Task accuracy &LLM rewriter &TopK selection  \\
    MOP \cite{mop} & Instruction induction & Fixed & Task accuracy & Mixture of experts & Region-based \newline joint search \\
    DSPY \cite{dspy} & Manually created +\newline Instruction induction &Variable & LLM feedback +\newline Task accuracy &Program Synthesis &TopK selection \\
    OPRO \cite{opro} & Manually created &Variable & LLM feedback +\newline Task accuracy &Metaprompt design &TopK selection  \\
    GATE \cite{gate} &Manually created &Variable &Human feedback &LLM rewriter &N/A  \\
    
\hline
\end{tabular}
}
\caption{Comparison of some APO techniques under our framework (Tables \ref{tab:comparison_1},\ref{tab:comparison_2},\ref{tab:comparison_3} show full comparison)}
\label{tab:comparison_0}
\end{table*}

\section{Inference Evaluation and Feedback} 
The evaluation step helps identify promising prompt candidates in each iteration. Some methods also use LLM feedback on prompt-response pairs to help generate more prompt candidates. 
\label{evaluate}
\subsection{Numeric Score Feedback}
\label{score-feedback}
\subsubsection{Accuracy} 
\label{accuracy}
Using task-specific accuracy metrics is the most straightforward and widespread way of eliciting feedback, i.a., \cite{ape, claps, sprig, dsp}. Classification and MCQ-based QA tasks use exact accuracy, while code-related tasks measure execution accuracy. Text generation tasks (summarization, translation, creative writing) employ flexible metrics like BLEU-N, Rouge-N, Rouge-N-F1, or embedding-based measures such as BERTScore \cite{bert-score} \cite{instruction-induction, pace}.

\subsubsection{Reward-model Scores}\label{reward}
Given the limitations of rigid accuracy metrics, some approaches proposed using learned reward models to provide more nuanced evaluations of prompts-response pairs \cite{rlprompt, oirl, prewrite}. OIRL \cite{oirl} trained an XGBoost-based reward model that takes query-prompt embedding pairs as input and predicts whether the prompt will elicit correct answers from the language model and use it to select appropriate prompts for specific queries using a best-of-N strategy. 
DRPO \cite{drpo} follows an LLM-based reward modeling approach using both predefined and dynamic reward criteria. It first optimizes in-context learning examples $E$, and using that it optimizes the specific task prompt.  

\subsubsection{Entropy-based Scores} \label{entropy}
Entropy-based scores evaluate the entire output distribution induced by candidates, as opposed to a single inference instance. They are gradient-free but require access to the entire output probability distribution, something not usually possible with black-box LLMs. CLAPS \cite{claps} leverages the negative incremental cross-entropy of  $\pi_{(x_{i} \oplus v\in V)}$ v/s $\pi_{(x_{i})}$ to identify promising words $v \in V$ to add to the prompt. The topK words are then used as candidate tokens from which to construct candidate prompts. GRIPS \cite{grips} simply added an entropy term to the task-weighted accuracy $-\sum \pi_{\rho}(y) \, ln(\pi_{\rho}(y)) + \frac{1}{|T|}\sum \mathbf{1}(y=\hat{y})$ to prioritize output diversity in potential prompt candidates. 

\subsubsection{Negative Log-likelihood of Output}
\label{nll}
Some approaches like APE, GPS \cite{gps}, PACE \cite{pace} consider the negative log-likelihood (NLL) of token sequences under the target LLM, i.e., $-\log(\pi_{\rho} (y))$. This however requires the log-probabilities to be accessible during the decoding of each token, limiting its applicability. The NLL for ground truth one-hot token-sequence is equivalent to the cross-entropy.  

\subsection{LLM Feedback}
\label{llm-feedback}
A popular paradigm to augment or fully replace numeric scores is to use textual feedback generated by $LLM_{Evaluator}$ \cite{promptagent, adv-icl, sos}. It is versatile because it can evaluate both the response as well as the prompt input. It can directly aid the prompt rewriting process while being flexible to individual tasks as it only needs natural language instructions for general-purpose LLMs as opposed to task-specific handcrafting of metrics. A potential downside is the inference cost incurred due to an additional LLM call. All the LLM feedback approaches provide multiple feedback data and broadly fall into two categories - improving a single prompt candidate versus improving multiple prompt candidates (discussed below, examples in Appendix \ref{appendix:tab_llmaaj}). 

\subsubsection{Improving Single Candidate}
\label{single}
SCULPT \cite{sculpt} introduces a systematic method for tuning long, unstructured prompts by employing a \textbf{hierarchical tree structure} and two-step feedback loops - preliminary assessment and error assessment - to evaluate and correct prompts before and after execution. The feedback updates the hierarchical prompt tree which is then back-synthesized into a new prompt candidate. PACE \cite{pace} applies an \textbf{actor-critic} editing framework to the prompt refinement process itself, allowing for more dynamic and adaptive adjustments.
Overcoming the limitations of optimizing a single metric, CRISPO \cite{crispo} adopts a \textbf{multi-aspect critique-suggestion} meta-prompt to highlight flaws in the generated response across multiple dimensions such as style, precision, and content alignment. Thereafter it leverages detailed, aspect-specific feedback and iteratively updates the prompts. Autohint \cite{hint} summarizes feedback for multiple incorrect inferences via \textbf{hints} to instill improvements into a single prompt candidate.

\subsubsection{Improving Multiple Candidates}
\label{multiple}
\ignore{\textbf{Textgradients:}} ProTeGi \cite{protegi} and TextGrad \cite{textgrad} leverage \textbf{textual “gradients”} to guide the discrete prompt optimization procedure, very similar to the gradient-descent style of continuous prompt optimization approaches. Different from continuous gradient-descent, ProTeGi sampled multiple “gradients” i.e. directions of improvement, and each such “gradient” is used to generate several prompt candidates for evaluation in the next iteration. PromptAgent \cite{promptagent} similarly used an error collection approach to emulate expert-written prompts that consisted of clear sections like “Task description”, “Domain Knowledge”, “Solution Guidance”, “Exception Handling”, “Output Formatting”. \ignore{\textbf{Ensemble learning}:} PREFER \cite{prefer} utilizes a feedback-reflect-refine cycle to aggregate feedback into multiple prompts in an \textbf{ensemble} to improve the model's ability to generalize across various tasks. \ignore{\textbf{Safety score:}} Survival of the Safest (SOS) \cite{sos} added \textbf{safety-score} into a multi-objective prompt optimization framework that used an interleaved strategy to balance performance and security in LLMs simultaneously. To avoid accidentally damaging well-functioning prompts, StraGo \cite{strago} summarized strategic guidance based on both correct and incorrect predictions as feedback. 



\subsection{Human-feedback}
\label{human}
A few works also incorporate human feedback, either during compile-time or inference-time in the prompt construction / optimization process. \citet{gate} proposed “Generative Active Task Elicitation” to better capture human preferences. It prompts a language model to interactively ask questions and infer human preferences conditioned on the history of free-form interaction. \citet{bpo-cheng} trained a smaller LLM to optimize input prompts based on user preference feedback, achieving up to 22\% increase in win rates for ChatGPT and 10\% for GPT-4. PROMST \cite{promst} tackles the challenges of multi-step tasks by incorporating human-designed feedback rules and a learned heuristic model. APOHF \cite{apohf} focuses on optimizing prompts using only human preference feedback rather than numeric scores, employing a dueling bandits-inspired strategy to efficiently select prompt pairs for preference feedback, proving effective for tasks like text-to-image generation and response optimization. 

\section{Candidate Prompt Generation}
\label{candidate}
In this step, one or more candidate prompts are generated that are most likely to result in an improvement in a metric of interest $f \in F$. The approaches reviewed below range from simple rule-based edits (sec. \ref{heuristic}) to sophisticated agentic systems that combine with LLM-based evaluations (sec. \ref{llm-feedback}) and various filtering strategies (sec. \ref{filter}). 

\subsection{Heuristic-based Edits}
\label{heuristic}
Several works proposed heuristic-based mechanisms to make edits to intermediate prompt candidates to generate newer candidates. They range from edits at the word / phrase / sentence-level (either simple rule-based or LLM-generated), or metric-driven incremental search. While these strategies may not result in the most optimal solution, they help in making the discrete prompt optimization problem computationally tractable. 
\subsubsection{Monte Carlo Sampling} 
\label{mc}
ProTeGi \cite{protegi} uses Monte carlo sampling to explore combinatorial discrete solution spaces in an incremental fashion - it samples multiple textual gradients to use to generate prospective candidates, and spawns paraphrases as monte-carlo successors for evaluation.  PromptAgent \cite{promptagent} uses a tree-variant called Monte Carlo Tree Search (MCTS) which consists of 4 steps — Selection, Expansion, Simulation, and Backpropagation (also explained in Sec. \ref{filter}).  

\subsubsection{Genetic Algorithm} 
\label{genetic}
A significant line of work applies the well-studied genetic algorithms to make discrete edits to texts. The common recipe for several genetic algorithms is 1/ Mutate and 2/ Cross-over components from promising candidates. 
\textbf{Token mutations:} SPRIG \cite{sprig} and CLAPS perform token-level mutations. SPRIG uses a starting corpus of 300 components grouped into categories like COT, roles, styles, emotions, scenarios, and good properties. It performs add/rephrase/swap/delete, highlighting complementary strengths of optimizing system prompts alongside task-prompts (via methods like ProTeGi) to enhance accuracy across multiple diverse domains, languages, and tasks without needing repeated task-specific optimizations. 

\textbf{LLM-based mutation:} LMEA \cite{lmea}, SOS \cite{sos}, and StraGo \cite{strago} uses mutation prompts with LLMs to overcome the traditional complexity of designing tailored operators for cross-over / mutation. PromptBreeder \cite{prompt-breeder} advocates self-referential improvement of all prompts in the prompt optimization system - Direct Mutation of task prompts, Hypermutation of mutation prompts themselves, Lamarckian Mutation where prompts are reverse-engineered from successful examples (similar to Instruction Induction \citet{instruction-induction}, and finally Crossover and Shuffling to improve diversity of the prompt pool. EvoPrompt \cite{evoprompt} use Differential Evolution - where differences between existing prompts is incorporated to form new prompt candidates to overcome the problem of local optima. AELP \cite{aelp} also uses mutation operators to perform sentence-level edits in an iterative fashion. They include sentence-level histories of reward $\{(s_{t-1},s_t,r_t)\}$ in the mutation prompt in order to avoid local optima and accidentally returning to sub-optimal versions. GPS \cite{gps} used Back-translation, Sentence Continuation, and Cloze transformations to perform prompt mutation. PromptWizard \cite{prompt-wizard} proposed a pipeline combining several steps including iterative improvement, few shot example synthesis and selection, utilizing LLM’s reasoning capability to improve and validate the prompt, and finally an expert persona to ensure consistency of the style of generated prompts. 

\subsubsection{Word / Phrase Level Edits} 
\label{word}
Several word-edit approaches first identify "influential" tokens in the prompts. COPLE \cite{cople} argued that LLMs exhibit lexical sensitivity, showing that merely replacing a few words with their synonyms can yield significant improvements. First, “influential” tokens are identified where expected loss on dev-set $E_{D_{val}} [L(y,\hat{y})]$ drops the most after removing that token versus the original prompt, and then influential tokens are replaced using predictions from a Masked-Language Models. This token-replacement approach is also attractive as a standalone post-processing step for long prompts that are already optimized using other LLM-based approaches. 
GRIPS \cite{grips} argues that phrase level edition is an effective and interpretable method to optimize prompts, leveraging 4 basic edit operations -add, delete, paraphrase, and swap

\subsubsection{Vocabulary Pruning}
\label{prune}
Some works prune the vocabulary space $V$ to $V_{pruned}$ for decoding the next token for the optimized prompt $\rho^*$. CLAPS \cite{claps} argued that general search spaces are highly redundant and use K-means clustering to find word-clusters and retain top-2000 words closest to cluster centroids. BDPL \cite{bdpl} used pairwise mutual information (PMI) to retain top co-occuring ngrams for decoding. PIN \cite{pin} instead added regularization in the form of Tsallis-entropy (ideal for heavy-tailed distributions like natural language) for the RL training of a prompt generation network, to reduce the probability mass for unlikely tokens and improve interpetability. 

\subsection{Editing via Auxiliary Trained NN}
\label{nn}
Some approaches leverage a trained auxiliary neural network to edit the initial prompt for obtaining desired improvements. We include approaches where the finetuned network is different and smaller than the task network. 

\subsubsection{Reinforcement-learning}
\label{rl}
\textbf{Multi-objective Optimization} techniques \cite{morl-prompt} demonstrate superiority over simple reward averaging, particularly through volume-based methods that effectively balance competing objectives. Dynamic prompt modification strategies, introduced through \textbf{prompt rewriting} \cite{prewrite}, directional stimulus prompting \cite{directional-stimulus} and \textbf{test-time editing} \cite{tempera} solve the important goal of moving beyond static prompt generation. Prompt-OIRL \cite{oirl} also tackled test-time optimization objective by learning an \textbf{offline reward model} and subsequently using a best-of-N strategy to recommend the optimal prompt in a query-dependent fashion. BDPL \cite{bdpl} optimized discrete prompts using variance-reduced policy gradient algorithm to estimate gradients, allowing user devices to fine-tune tasks with limited API calls. 
\subsubsection{Finetuning LLMs} 
\label{ft}
BPO \cite{bpo-cheng} trains a smaller 7B model to align itself to task-performance on individual LLMs using reward-free alignment. FIPO \cite{fipo} trains a local model (7B - 13B) to perform prompt optimizations to preserve privacy and adapt to target models better leveraging both data diversification and strategic fine-tuning such as SFT, preference optimization, and iterative preference learning. 
\subsubsection{Generative Adversarial Networks} 
\label{gan}
\citet{adv-icl} framed the prompt optimization process in the GAN setting. The LLM generator takes question and the generation prompt to produce output. The (input, output) pairs are evaluated by an LLM powered discriminator, whose goal is to identify generated pairs from ground truth pairs. Both generator and the discriminator are jointly optimized using adversarial loss, by utilizing a prompt modifier LLM to rewrite their prompts. 

\subsection{Metaprompt Design}
\label{metaprompt}
PE2 \cite{pe2} argued that previous works under-explored meta-prompt search space. OPRO \citep{opro} proposes a meta-prompt design (see Appendix \ref{appendix:metaprompt}) which includes the optimization problem description in natural language and previously generated solutions (multiple solutions per stage for diversity) and scores alongisde the meta-instruction for prompt refinement. 
DAPO \cite{dapo} utilizes a well-designed meta-instruction to guide the LLM in generating high-quality and structured initial prompts (contain task-specific info, e.g. task type and description, output format and constraints, reasoning process, professional tips) by observing given input-output exemplars. Then, DAPO iteratively optimizes the prompts at the sentence level, leveraging previous tuning experience to expand prompt candidates.

\subsection{Coverage-based}
\label{coverage}
Some approaches seek to "cover" the entire problem space - either within a single prompt, or using multiple prompts working individually or in an ensemble during inference. 

\subsubsection{Single Prompt-expansion} 
\label{expand}
AMPO \cite{ampo} uses LLM feedback to enumerate all the failure cases based on the evaluation-set $D_{val}$ and then enlists each of them in the meta-instruction in an if-then-else format using 3 modules - 1/ Pattern Recognition, 2/ Branch Adjustment, and 3/ Branch Pruning to decide whether to enhance existing branches, or to grow new branches. Similarly, UNIPROMPT focused on explicitly ensuring that various semantic facets of a task get represented in the final prompt. It designs a human-like (manual) prompt engineering approach (UniPrompt) with two stages: a) task facets initialization using background knowledge, and b) refinement using examples. 

\subsubsection{Mixture of Experts}
\label{moe}
\citet{mop} introduced the Mixture-of-Expert-Prompts where each expert is a task-prompt to be used for specialized inference. MOP first clusters all demonstrations using K-means clustering. Then, the Region-based Joint Search (RBJS) (sec.\ref{rbjs}) algorithm generates the appropriate instruction for each exemplar-cluster via instruction induction (sec.\ref{induction}) based on a mix of in-cluster and out-of-cluster demonstrations to cover “blind-spots”. During inference, a single expert prompt is invoked whose cluster centroid $\mu_c$ is closest to the instance-embedding $\arg \min_{_{C}} ||\phi(x_i) - \mu_c||_2$. 

\subsubsection{Ensemble Methods}
\label{ensemble}
PromptBoosting \cite{prompt-boosting}, BoostedPrompting \cite{boosted-prompting}, PREFER \cite{prefer}, etc. are ensemble methods that invoke multiple prompts during inference and combine them to generate the final output $\hat{y} = y_0 + \Sigma_m \beta_i y_i$. GPO \cite{gpo} also uses labeled source data to generate an ensemble of prompts, which are applied to unlabeled target data to generate output through majority voting. 

\subsection{Program Synthesis}
\label{program}
Program-synthesis based approaches transform LLM pipelines into structured, modular components that can be systematically optimized and composed\ignore{, rather than using static prompts}. These optimization techniques iteratively refine instructions and demonstrations for each module to improve the entire pipeline's performance, \ignore{using program-aware methods and meta-optimization procedures.}
DSP \cite{dsp} introduces a three-stage framework for retrieval-augmented inference: Demonstrate (generates task-specific demonstrations), Search (retrieves relevant information), and Predict (combines retrieved info with demonstrations). DSPY \cite{dspy} transforms LLM pipelines into text transformation graphs - introducing parameterized models, learning through demonstrations, and a compiler that optimizes pipelines. DLN \cite{dln} similarly considers chained LLM calls as stacked deep language networks performing variational inference, where the learnable parameters for each layer are task-decomposed prompt templates. MIPRO \cite{mipro} automates the optimization of multi-stage language model programs by improving instructions and demonstrations for each module. SAMMO \cite{sammo} proposed symbolic prompt programming, representing prompts as directed-acyclic-graphs (DAG). A set of user-defined node mutation rules guide the mutation-search to find the optimal DAG\ignore{based on $E_{D_{val}}[f(\rho)]$}, which is then converted back to a prompt. 

\section{Filter and Retain Promising Prompts}
\label{filter}
In this step, promising prompt candidates are filtered for further optimization. 
\subsection{TopK Greedy Search}
\label{topk}
The simplest mechanism to iteratively search through prompt candidate sets is a greedy topK search where in each iteration of the optimization, the top-K best-performing candidates on mini-batch of data instances $D_{val}$ are retained for further iterations (e.g. - ProTeGi, AELP. This differs from beam-search which judges partial solutions’ 
based on the reward for the entire trajectory of prompt edits $r(\{\rho_{1}^1, \rho_{2}^1, \ldots,\rho_{t}^1\})$.
\subsection{Upper Confidence Bound and Variants}
\label{ucb}
Relying on a single static evaluation dataset can lead to biases in the selection procedure and finally suboptimal solutions. ProTeGi, SPRIG, \textit{inter alia}, cast the candidate prompt selection problem as that of bandit search - identifying the most suitable arm (prompt candidate) operating on a fixed computation budget. They use the Upper Confidence Bounds (UCB, Algorithm \ref{algo:ucb}) which balances exploration with exploitation. In each iteration of prompt optimization, they sample a different evaluation dataset $D_{sample} \in D_{val}$, and maintain a moving estimate of the optimality of each arm (i.e. prompt). In each iteration, the playout filters top-B prompt candidates with the greatest score for further exploration. PromptAgent uses a variation of UCB called UCB for Trees (UCT) which are used in the setting of contextual bandits (i.e. the action-space and the reward function is state-dependent). AELP \cite{aelp} used a modification called Linear UCB \cite{lin-ucb} which uses a closed form linear estimate based on the reward trajectories of previously sampled edits as well as prompt embedding $\phi(s)$ to select the next best-arm. 
\subsection{Region-based Joint Search}
\label{rbjs}
MOP \cite{mop} proposes a Mixture-of-Expert-Prompts performing prompt optimization for each expert individual. Once C exemplar-clusters are identified, the RBJS search first samples examples  $D_{exemplars} \in D_C \cup D \setminus D_C$, and then uses APE to induct and optimize each expert instruction.  
\subsection{Metaheuristic Ensemble}
\label{metaheuristic}
PLUM \cite{plum} library offered a meta-heuristic ensemble of different search algorithms like Hill climbing, Simulated Annealing, Genetic Algorithms, Tabu Search, and Harmony Search. 

\section{Iteration Depth}
\label{depth}
\subsection{Fixed Steps}
\label{fixed}
Most approaches choose to carry out the prompt optimization for a fixed number of steps N. 
\subsection{Variable number of steps}
\label{variable}
GRIPS \cite{grips} concludes search when successive iterations with negative gains breach a patience parameter, whereas PromptAgent concluded APO when $r_t \leq \epsilon_{min} \vee r_t \geq \epsilon_{max}$. 
\section{Theoretical Perspectives}
\subsection{Upper Bound of Improvement from APO}
\label{ub}
\ignore{While there are theoretical explorations behind In-Context Learning \cite{icl-survey}, there is a relative dearth of such theoretical understanding of Instruction Optimization techniques. It is useful to understand prompt optimization from a theoretical perspective to establish the lower and upper bound of improvement attainable under a given prompt optimizer.} AlignPro \cite{align-pro} establishes an upper bound on the gains realizable from discrete prompt optimization under a given prompt optimizer and also a suboptimality-gap w.r.t. RLHF-optimal policy $\pi^*$, while a lower bound is left unexplored. 
\subsection{Other Related Perspectives}
\citet{control} proposed a control theoretic framework to establish bounds on the set of reachable LLM-outputs for self-attention in terms of the singular values of its weight matrices. \citet{universality} showed the existence of a strong transformer \ignore{with a prompt} that can approximate any sequence-to-sequence Lipschitz function. They also showed the existence of “difficult” datasets that depth-limited transformers could not commit to memory.

\section{Challenges and Future Directions}
\subsection{Task-agnostic APO}

All the surveyed APO methods assume that the task type $T$ is known beforehand; additionally offline APO methods also require an evaluation set $D_{val}$, something not explicitly available in production settings. Barring a few tasks covered by \citet{gate, oirl, tempera, pin}, inference-time optimization of multiple unknown tasks is underexplored. \ignore{MOP solves a related yet incomplete setting where the data instance is routed to the most suitable expert, but doesn’t explore unseen task-prompts $\rho_{unseen} \in T^*, T \subseteq T^* , \rho_{unseen}:= I_{unseen} \oplus E_{unseen} \oplus \tau$.} More robust evaluations are needed for task-agnostic APO systems combining seen and unseen tasks. 

\subsection{Unclear Mechanisms}
\citet{propane} showed that prompts have so-called 'evil twins' that are uninterpretable yet recover some of the performance of gold-standard prompts. \citet{random-separators} showed that rare gibberish strings can serve as competitive delimiters $\tau$ in prompts. \citet{abo} showed that self-reflection by LLMs can suffer from incorrect error identification, prior biases, semantic invalidity, leading to failure in yielding improved prompts. More studies are needed to better uncover the mechanisms of prompt optimization. 
\subsection{APO for System Prompts / Agents}
Although SPRIG explored optimizing system prompts in chat-style settings, scalability remains a challenge - optimizing system prompts required a predefined corpus and close to 60 hours whereas Protegi only needed \~10 minutes per task. Similarly, optimizing prompts for several components in an agentic system in a concurrent fashion poses an exciting direction for future research.  

\subsection{Multimodal APO}
Recently, textual prompt optimization has expanded to multimodal domains: text-to-image~\cite{liu2024language, manas2024improving, liu2024you}, text-to-video~\cite{ji2024prompt}, text-to-audio~\cite{huang2023make}, and text-image alignment models like CLIP~\cite{du2024ipo, mirza2024glov}. Beyond textual prompts, ~\citet{huang2023make} explore optimizing multimodal inputs, such as images, to elicit better responses from large multimodal models. However, the interplay between modalities in prompt optimization remains underexplored. Future research could develop APO frameworks to jointly optimize multimodal prompts (eg - remove background noise from audio, add visual markers to videos, etc.) to fully leverage their synergies.

\section{Conclusion}
\label{sec:conclusion}
In this paper, we provide a comprehensive fine-grained review of existing APO techniques and identified key areas for future growth. It is our aim to spur future research spawning from our survey. 

\section{Limitations}
\label{sec:limitation}
While we attempted to cover all qualifying papers, it is possible that we may have unintentionally missed out on some relevant papers. We also mention some of the papers that were excluded in this survey with specific reasons in section \ref{excluded}. Also, we realize that fitting varied research works into a single unifying framework might risk broad categorizations for some papers, or skipping some characteristics for others (e.g. Tempera \cite{tempera} consists of both RL-based and word/phrase-level editing techniques, applied to both instructions and exemplars). In such cases, we categorize a paper based on its most salient features. Another challenge is that when presenting a survey paper under 8 pages, we had to make tradeoffs and only retain content in the main body that was deemed most necessary. This resulted in having to relegate a core contribution (Tables \ref{tab:comparison_1},\ref{tab:comparison_2},\ref{tab:comparison_3}) which contained a rigorous comparison of all the surveyed papers into the appendix. We have attempted our best to strike the right balance between specificity and brevity to present a novel framework. We also provide copious references to interested researchers for further reading. 

\onecolumn
\section{Appendix}
\label{sec:appendix}

\subsection{Notation}
\label{notation}
We now define the notation of key terms and expressions used throughout the paper. 
\begin{enumerate}
\itemsep0em 
\item $T$ = Task type, $I$= Task instruction, $E={(xi,yi)}_{i=1}^e$ Few shot demonstrations in the prompt, $\tau$= Template delimiters, z = CoT recipe for a task-instance, $z_i \in I_i$
\item{$M_{task}$ target model, $M_{APO}$ APO system}
\item $\rho=concat([s_1,s_2,\ldots,s_m])=concat(I,\tau,E)$ Prompt composed of m sentences, which comprise of Instruction, template delimiters and few-shot demonstrations. 
\item $D=\{(x_i,y_i)\}_{i=1}^m$ collection of m input-output pairs. $D_{val}$ is the validation set used to validate prompt performance, $D_{train}$ is the training set used to finetune the language model(Reprompting). 
\item $\{f_1,f_2,\ldots\} \in F$ metric function upon which to evaluate task-prompt performance
\item $r:S\times A\rightarrow R$= reward model score, where S is the state-space and A is the action-space
\item $|V|$ = length of vocabulary
\item $\phi:S\in V_* \rightarrow R_d$ embedding function which takes in a sentence generated as a finite sequence of tokens belonging to a vocabulary V, and generating a floating point array representation of dimension d
\item $\rho_*=argmax_{\rho \in V_*}E_{D_{val}}[f_i(\rho)]$ The best performing prompt based on the metric score on validation set 
\item $k$ = number of candidates for top-K search, $B$ = Beam width for beam search, $N$ = number of iterations for search
\item $C$ = number of experts in a Mixture of Experts approach (MOP), $\mu_C$= cluster centroid of cluster C (MOP). 
\item $LLM_{target}$= target model which will be used for inference, $LLM_{rewriter}$= rewriter model which will be used for rewriter, $LLM_{evaluator}$= evaluator model which provides the LLM feedback to prompts / responses or both
\item $\lambda$ with subscripts to denote different latency types: 
$\lambda_t$ = Total training cost/latency, including all offline costs for data collection, preprocessing, and model fine‐tuning, $\lambda_i$ = per-example inference latency, $\lambda_m$ = MLM inference latency per-example

\end{enumerate}

\subsection{Excluded works}
\label{excluded}
\textbf{FedBPT} \cite{fedbpt} used federated learning to update soft prompts and not discrete tokens. \textbf{Deliberate-then-generate} \cite{DTG} randomly sampled arbitrary noisy inference and prompted the task LLM to deliberate on the wrong inference, while \textbf{Reflexion} \cite{reflexion} agents maintain an episodic buffer of past deliberations. Neither method optimizes the input prompt. \textbf{AutoPrompt} \cite{autoprompt} required gradient access to the task LLM and therefore doesn't remain blackbox. 
\clearpage
\subsection{UCB based selection algorithm}
\begin{small}
\begin{algorithm}
\caption{$Select(\cdot)$ with UCB Bandits}
\label{algo:ucb}
\begin{algorithmic}[1]
\Require $n$ prompts $\rho_1, ..., \rho_n$, dataset $\mathcal{D}_{val}$, $T$ time steps, metric function $m$
\State Initialize: $N_t(\rho_i) \gets 0$ for all $i = 1, \dots, n$
\State Initialize: $Q_t(\rho_i) \gets 0$ for all $i = 1, \dots, n$
\For{$t =1, \dots, T$}
    \State Sample uniformly $\mathcal{D}_{sample} \subset \mathcal{D}_{val}$
    \State $\rho_i \leftarrow
\arg\max_\rho \left\{\frac{Q_t(\rho)}{N_t(\rho_i)} + c \sqrt{\frac{\log t}{N_t(\rho)}}\right\}$
    \State Observe reward $r_{i,t} = m(\rho_i, \mathcal{D}_{sample})$
    \State $N_t(\rho_i) \gets N_t(\rho_i) + \vert \mathcal{D}_{sample} \vert$
    \State $Q_t(\rho_i) \gets Q_t(\rho_i) + r_{i, t}$
\EndFor
\State \Return $SelectTop_b(Q_T/N_T)$
\end{algorithmic}
\end{algorithm}
\end{small}

\section{Comparison of different approaches + Tasks}
\subsection{Comparison}
Below we offer a comprehensive comparison of all the surveyed methods against our framework, covering the following aspects
\begin{enumerate}
\itemsep0em 
\item \textbf{Seed instructions}
\item \textbf{Inference evaluation}
\item \textbf{Candidate generation} 
\item \textbf{Search+filter strategy} 
\item \textbf{Iteration depth} 
\item \textbf{Optimization time complexity}
\item \textbf{Prompt generation model}
\item \textbf{Target models}
\end{enumerate}


\begin{sidewaystable*}
\rowcolors{2}{gray!15}{white} 
\resizebox{1\textwidth}{!}
{\begin{tabular}{|p{3cm}p{4cm}p{2.5cm}p{3.5cm}p{3.5cm}p{4cm}p{3cm}p{3cm}p{3cm}p{5cm}|} 
\hline
\textbf{SNo.} & \textbf{Method} & \textbf{Seed instructions} & \textbf{Inference evaluation} & \textbf{Candidate generation} & \textbf{Search+filter strategy}  & \textbf{Iteration depth}& \textbf{Optimization time complexity} & \textbf{Prompt generation model}& \textbf{Target models}\\
\hline
1 & GPS \cite{gps} & Manually created & Task accuracy & Genetic Algorithm:\newline Back translation, Cloze, \newline Sentence continuation & Metaheuristic ensemble & Fixed & $O(T * N * k * \lambda_i)$ &  & T0 \\
2 & GRIPS \cite{grips} & Manually created & Entropy-based score+\newline Task accuracy & Phrase level \newline add/remove/swap/paraphrase & TopK selection & Fixed & $O(k * N * |D_{val}| * B)$ & PEGASUS paraphrase model & InstructGPT \\
3 & Instruction induction \newline \cite{instruction-induction} & Instruction induction & Accuracy + \newline BERTScore & LLM-rewriter &  & Fixed & $O(|\rho| * \lambda_i)$ & InstructGPT, GPT-3 & InstructGPT, GPT-3 \\
4 & RLPrompt \cite{rlprompt} & Manually created & Task accuracy + \newline Reward model score & RL-based trained NN & TopK selection & Fixed & $O(N * \rho * |V| * \lambda_i)$ & RoBERTa-large \newline Reward model-DistilBERT & 1/ BERT, 2/ GPT-2 \\
5 & TEMPERA \cite{tempera} & Manually created & Task accuracy & RL-trained NN &  & Fixed & $O(N * k * |V| * C)$ & RoBERTa-large & RoBERTa-large \\
6 & AELP \cite{aelp} & Manually created & Task accuracy & Genetic algorithm: \newline LLM-mutator & Beam search & Fixed & $O(N * \rho * k * |D| * \lambda_i)$ & PaLM 2-L & PaLM text-bison \\
7 & APE \cite{ape} & Instruction induction & Task accuracy & No new candidates & TopK selection & Fixed & $O(N * k * |D_{val}| * \lambda_i)$ & InstructGPT, GPT-3, T5, \newline InsertGPT & InstructGPT, GPT-3 \\
8 & AutoHint \cite{hint} & Manually created & Task accuracy + \newline LLM-feedback & LLM rewriter & TopK selection & Fixed & $O(T * |D| * \lambda_i)$ &  & GPT-4 \\
9 & BDPL \cite{bdpl} & Manually created & Task accuracy & RL-trained NN & TopK selection & Variable & $O(N * k * \lambda_i)$ & RoBERTa, GPT-3 & RoBERTa, GPT-3 \\
10 & Boosted Prompting \newline \cite{boosted-prompting} & Instruction-induction & Task accuracy & Ensemble based method & TopK selection & Variable & $O(N * k * \lambda_i)$ & text-curie-001, text-curie-003, GPT-3.5,\newline  code-davinci-002 & text-curie-001, text-curie-003, \newline GPT-3.5, code-davinci-002 \\
11 & BPO \cite{bpo-cheng} & Manually created & LLMaaJ (pairwise) & Finetuned LLMs &NA &NA & $O(\lambda_t + |D_{val}| * \lambda_i)$ & Llama2-7b-chat & Vicuna-7b-v1.3, \newline vicuna-13b-v1.3, llama-1-7b, \newline llama-1-13b \\
12 & CLAPS \cite{claps} & Manually created & Entropy-based score+\newline Task accuracy & Genetic Algorithm: \newline Mutation + Crossover & TopK selection & Variable & $O(N * k * |V| * \lambda_i)$ & Flan-T5 & Flan-T5 large and base \\
13 & Directional-stimulus \cite{directional-stimulus} & Manually created & BLEU, BERTScore & RL-trained NN &  & Variable & $O(\lambda_t)$ & T5, GPT-2 & ChatGPT, Codex, InstructGPT \\
14 & DLN \cite{dln} & Manually created & Task accuracy + NLL & LLM mutator & TopK selection & Fixed & $O(N * k * |D_{train}|)$ & GPT-3 (text-davinci-003), GPT-4 & GPT-3 (text-davinci-003), GPT-4 \\
15 & DSP \cite{dsp} & Instruction induction & Task accuracy & Program Synthesis & TopK selection & Fixed & $O(N * k * \lambda_i)$ & GPT-3.5 & LM: GPT-3.5, \newline Retrieval: ColBERTv2 \\
16 & DSPy \cite{dspy} & Manually created + \newline Instruction Induction & Task accuracy + \newline LLM-feedback & Program Synthesis & TopK selection & Variable & $O(N * k * B * \lambda_i)$ &  &  \\
17 & GATE \cite{gate} & Manually created & Human feedback & LLM rewriter &  & Open-ended & $O(N * (\lambda_m + |D_{val}| * \lambda_i))$ & GPT-4 & GPT-4 \\
18 & GPO \cite{gpo} & Instruction induction & Task-Accuracy and F1 & Metaprompt-design & TopK selection &  & $O(N * C * |V| * B * E)$ & gpt-3.5-turbo-0301 & gpt-3.5-turbo-0301 \\
19 & PACE \cite{pace} & Manually created & NLL + Task accuracy - \newline BLEU and BERTScore & LLM-rewriter & TopK selection & < 3 & $O(N * |\rho| * |D_{val}|)$ & gpt-3.5-turbo (0301) & text-davinci-002, \newline text-davinci-003,\newline  (gpt-3.5-turbo), GPT-4 \\
20 & PREFER \cite{prefer} & Manually created & Task accuracy & LLM-rewriter + \newline Ensemble method & TopK selection & Fixed & $O(N * |\rho| * |D_{val}|)$ & ChatGPT & ChatGPT \\
21 & Promptagent \cite{promptagent} & Manually created & Task accuracy + \newline LLM-feedback & LLM rewriter & UCT-based bandit-search & Fixed & $O(N * k * \lambda_i)$ & GPT-4 & GPT-3.5, GPT-4, PaLM-2 \\
\hline
\end{tabular}}
\caption{Comparison of all APO techniques based on our framework}
\label{tab:comparison_1}
\end{sidewaystable*}

\begin{sidewaystable*}
\rowcolors{2}{gray!15}{white} 
\resizebox{1\textwidth}{!}
{\begin{tabular}{|p{3cm}p{4cm}p{2.5cm}p{3.5cm}p{3.5cm}p{4cm}p{3cm}p{3cm}p{3cm}p{5cm}|} 
\hline
\textbf{SNo.} & \textbf{Method} & \textbf{Seed instructions} & \textbf{Inference evaluation} & \textbf{Candidate generation} & \textbf{Search+filter strategy}  & \textbf{Iteration depth}& \textbf{Optimization time complexity} & \textbf{Prompt generation model}& \textbf{Target models}\\
\hline
22 & Promptboosting \cite{prompt-boosting} & Instruction-induction & Accuracy, F1 Score & Ensemble based method & Beam-search & Early Stopping & $O(\lambda_m)$ & T5 & RoBERTa-large \\
23 & Promptbreeder \cite{prompt-breeder} & Manually created & LLM Feedback + \newline Task accuracy & Genetic Algorithm:\newline Mutate + Crossover\newline (LLM-edits) & Metaheuristic Ensemble  & Fixed & $O(\rho * N * |V| * \lambda_i)$ & text-davinci-003, PaLM 2-L & text-davinci-003, PaLM 2-L \\
24 & ProTeGi \cite{protegi} & Manually created & Task accuracy + \newline LLM-feedback & LLM rewriter & UCT-based bandit-search & Fixed & $O(N * C * |D_{val}| * \lambda_i)$ & GPT-3.5-Turbo & GPT-3.5-turbo \\
25 & Random separators \cite{random-separators} & Manually created & Task accuracy & LLM-rewriter & TopK selection & Fixed steps & $O(N * k * \lambda_)$ & GPT2 Large, GPT2 XL, \newline Mistral 7B, Mistral 7B Instruct, \newline Llama-Alpaca 7B, Llama2 7B. \newline Llama2 7B Chat, ChatGPT & GPT2 Large, GPT2 XL, \newline Mistral 7B, Mistral 7B Instruct, \newline Llama-Alpaca 7B, Llama2 7B. \newline Llama2 7B Chat, ChatGPT \\
26 & ABO \cite{abo} & Manually created + \newline Instruction Induction & Task accuracy + \newline LLM-feedback & LLM-rewriter & TopK selection & Fixed Steps & $O(B * N * \lambda_i)$ & GPT-4 & GPT-3.5-Turbo, Llama-2-70B-chat \\
27 & Adv-ICL \cite{adv-icl} & Manually created & LLM Feedback & LLM-rewriter & Top-1 selection & Fixed & $O(N * k * \lambda_i)$ & text-davinci-002, vicuna,\newline  ChatGPT & text-davinci-002, vicuna, ChatGPT \\
28 & AMPO \cite{ampo} & Manually created & Task accuracy + \newline F1 score & Coverage-based & TopK selection & Variable & $O(N * C * \lambda_i)$ & GPT-4-turbo & GPT-4-turbo \\
29 & APEER \cite{apeer} & Manually created & Task accuracy-nDCG & Feedback + preference \newline optimization &  & Used 3 epochs & $O(N * |\rho| * |D_val|)$ &  & GPT4, GPT3.5, Llama3, Qwen2 \\
30 & APOHF \cite{apohf} & Manually created & Task accuracy + \newline Human feedback & LLM rewriter & Linear UCB & Fixed & $O(N * T)$ & ChatGPT & DALLE-3, ChatGPT \\
31 & BATPrompt \cite{bat-prompt} & Manually created & Task accuracy + \newline LLM-feedback & LLM rewriter & TopK selection & Fixed & $O(N * |D| * |\rho| * \lambda_i)$ & GPT-3.5-turbo & GPT-3.5-turbo, \newline GPT-4o-mini, Llama2-7b \\
32 & COPLE \cite{cople} & Manually created & Task accuracy & Token edits using \newline MLM &  & Variable & $O(N * |I| * k * |D_val| * \lambda_i)$ & RoBERTa \newline (filling masked tokens) & Llama-2-7B-chat , \newline  Mistral-7B-Instruct-v0.1,\newline  ChatGPT (gpt-3.5-turbo-0125) \\
33 & CRISPO \cite{crispo} & Manually created & LLM feedback + \newline ROUGE-1/2/L F-measure, \newline AlignScore & LLM rewriter & TOP-K greedy search & Fixed & $O(N * k * (|D_{train}| * \lambda_i +\lambda_m))$ & Claude Instant, \newline Claude 3 Sonnet, \newline Mistral 7B, Llama3 8B & Claude Instant, \newline Claude 3 Sonnet, \newline Mistral 7B, Llama3 8B \\
34 & DAPO \cite{dapo} & Manually created & Task accuracy & LLM-rewriter & Top-1 selection & Fixed & $O(N * k * \lambda_i)$ & GPT-3.5-Turbo, Baichuan2, \newline GPT-4 & GPT-3.5-Turbo, \newline Baichuan2, GPT-4 \\
35 & DRPO \cite{drpo} & Manually created & Reward model score + \newline LLM Feedback & LLM rewriter & Beam search & Fixed & $O(B * k * N)$ & Mistral 7b, Mistral 7b (Instruct), \newline Llama 2 70b, Llama 2 70b (chat), \newline Llama 3 8b, Llama 3 8b (Instruct), \newline gpt-3.5-turbo & Mistral 7b, Mistral 7b (Instruct), \newline Llama 2 70b, Llama 2 70b (chat), \newline Llama 3 8b, Llama 3 8b (Instruct), \newline gpt-3.5-turbo \\
36 & EVOPROMPT \cite{evoprompt} & Manually created + \newline Instruction Induction & Task Accuracy + \newline ROUGUE+ SARI & Genetic Algorithm:\newline Mutation operators+\newline Crossover & Metaheuristic ensemble & Early Stopping & $O(N * k * T * \lambda_i)$ &  & Alpaca-7b, GPT-3.5 \\
37 & FIPO \cite{fipo} & Manually created & Task accuracy & Finetuned LLMs &  &  & $O(\lambda_t + |D_val| * \lambda_i)
)$ & Tulu-13B, Tulu-70B & Llama2-7B, Tulu2-13B, \newline Baichuan2-13B \\
\hline
\end{tabular}}
\caption{Comparison of all APO techniques based on our framework}
\label{tab:comparison_2}
\end{sidewaystable*}

\begin{sidewaystable*}
\rowcolors{2}{gray!15}{white} 
\resizebox{1\textwidth}{!}
{\begin{tabular}{|p{3cm}p{4cm}p{2.5cm}p{3.5cm}p{3.5cm}p{4cm}p{3cm}p{3cm}p{3cm}p{5cm}|} 
\hline
\textbf{SNo.} & \textbf{Method} & \textbf{Seed instructions} & \textbf{Inference evaluation} & \textbf{Candidate generation} & \textbf{Search+filter strategy}  & \textbf{Iteration depth}& \textbf{Optimization time complexity} & \textbf{Prompt generation model}& \textbf{Target models}\\
\hline

38 & LMEA \cite{lmea} & Manually created & Numeric Score-based & Genetic Algorithm:\newline Mutate + Crossover\newline (LLM-edits) & TopK selection & Fixed & $O(N * k * \lambda_i)$ &  & GPT-3.5-turbo-0613 \\
39 & MIPRO \cite{mipro} & Manually created & Task accuracy & Program Synthesis & TopK selection & Fixed & $O(N * |D_{val}| * k * \lambda_i)$ & GPT-3.5 (proposer LM) & Llama-3-8B (task LM) \\
40 & MOP \cite{mop} & Instruction induction & Task Accuracy & APE for each cluster & TopK selection & Fixed steps per-cluster & $O(C * N * |D_{val}|)$ & GPT-3.5-Turbo & GPT-3.5-Turbo \\
41 & MORL-Prompt \cite{morl-prompt} & Manually created & Task accuracy + \newline Reward score & RL-based trained NN &  & Fixed & $O(N * C * |V| * k)$ & distilGPT-2 & GPT-2 (style transfer),\newline  flan-T5-small (translation) \\
42 & OIRL \cite{oirl} & Manually created & Task accuracy + \newline Reward model score & LLM rewriter &  &  & $O(|D_{train}| * \rho * \lambda_i + \lambda_t + |D_{val}|* \lambda_i
)$ & GPT4 & Llama2-7B-chat, \newline Tigerbot-13B-chat, gpt3.5-turbo \\
43 & OPRO \cite{opro} & Manually created & Task accuracy + \newline LLM-feedback & Metaprompt design & TopK selection & Variable & $O(N * k * \lambda_i)$ & PaLM 2-L, text-bison,\newline  gpt-3.5-turbo and GPT-4 & PaLM family models \\
44 & PE2 \cite{pe2} & Manually created + \newline Instruction Induction & Task accuracy + \newline LLM-feedback & Metaprompt design & TopK selection & Fixed & $O(N * k * \lambda_i)$ & GPT-4 & text-davinci-003 \\
45 & PIN \cite{pin} & Manually created & Task accuracy & RL-trained LLM & TopK selection & Fixed & $O(N * |V| * \lambda_i * C)$ & OPT & RoBERTa-large (classification), \newline OPT models (others) \\
46 & PLUM \cite{plum} & Manually created & Task accuracy & Genetic Algorithm: \newline Mutate + crossover & Metaheuristics & Fixed steps & $O(N * C * k * \lambda_i)$ & GPT-3-babbage & GPT-3-babbage \\
47 & PRewrite \cite{prewrite} & Manually created & Task accuracy + \newline Reward model score & RL-trained LLM & TopK selection & Fixed & $O(N * C * \lambda_i * |V|)$ & PaLM 2-S & PaLM 2-L \\
48 & PROMPTWIZARD \cite{prompt-wizard} & Manually created & Task accuracy + \newline LLM-feedback & Genetic Algorithm: \newline Mutate + Crossover\newline (LLM-edits) & TopK selection & Fixed & $O(N * C * \lambda_i)$ & GPT3.5/GPT4 & GPT3.5/GPT4/Llama-70B \\
49 & PROMST \cite{promst} & Manually created & Task accuracy + \newline Human feedback & LLM rewriter & TopK selection & Fixed & $O(N * k * \lambda_i)$ & GPT-4 & GPT-3.5, GPT-4 \\
50 & Reprompting \cite{reprompting} & LLM generated CoT process. & Task accuracy & LLM-rewriter & Rejection sampling \newline with exploration & Fixed or until convergence & $O(N * k * |\rho|)$ & gpt-3.5-turbo, textdavinci-003 & gpt-3.5-turbo, textdavinci-003 \\
51 & SAMMO \cite{sammo} & Manually created & Task accuracy & Program synthesis & TopK selection & Fixed & $O(N * k * \lambda_i)$ &  & Mixtral7x8B, Llama-2 70B, \newline GPT3.5, GPT4 \\
52 & SCULPT \cite{sculpt} & Instruction induction \newline on task-README & Task accuracy + \newline LLM-feedback & LLM-rewriter & UCB bandit search & Fixed & $O(N * k * |\rho| * |D_val|)$ & GPT-4o & GPT-4o and Llama3.1-8B \\
53 & SOS \cite{sos} & Manually created & Task accuracy + \newline LLM-feedback & LLM-mutator & TopK selection & Fixed & $O(N * C * k * \lambda_i)$ & GPT-3.5-turbo, Llama3-8B, \newline Mistral-7B & GPT-3.5-turbo, Llama3-8B, \newline Mistral-7B \\
54 & SPRIG \cite{sprig} & Manually created & Task accuracy & Genetic Algorithm:\newline Mutate + Crossover (tokens) & Beam-search & Fixed & $O(N * B * T * k * \lambda_i)$ & tuner007/pegasus\_paraphrase & Llama 3.1-8B Instruct, \newline Mistral Nemo Instruct 2407, \newline Qwen 2.5-7B Instruct, \newline Llama 70B, Qwen 2.5-72B, \newline Mistral Large 2407. \\
55 & StraGo \cite{strago} & Manually created & Task accuracy + \newline LLM-feedback & Genetic Algorithm:\newline Mutate + CrossOver (tokens) & Bandit Search (UCB) & Early Stopping & $O(N * k * T * \lambda_i)$ & GPT-4 & GPT-3.5-turbo or GPT-4 \\
56 & TextGrad \cite{textgrad} & Manually created & Task accuracy + \newline LLM-feedback & LLM rewriter &  & Variable & $O(N * |D_{val}| * \lambda_i)$ &  & GPT-3.5, GPT-4o \\
57 & UNIPROMPT \cite{uniprompt} & Manually created + \newline Instruction Induction & Task accuracy + \newline LLM-feedback & LLM-rewriter & Beam Search & Early Stopping & $O(N * k * \lambda_i)$ & Fine-tuned Llama2-13B & GPT-3.5 \\
\hline
\end{tabular}}
\caption{Comparison of all APO techniques based on our framework}
\label{tab:comparison_3}
\end{sidewaystable*}

\label{full-table}
\clearpage
\subsection{Evaluation tasks and datasets}
Below we describe the different datasets and tasks that each method was evaluated on. 
\begin{table*}[!ht]
\rowcolors{2}{gray!15}{white} 
\resizebox{\textwidth}{!}{%
\begin{tabular}{|ccp{0.9\textwidth}|}%
\hline
\textbf{SNo.} & \textbf{Paper} & \textbf{Tasks}\\
\hline

1 & GPS \cite{gps} & 10 unseen tasks from the T0 benchmark, which span: \newline{} 1. Natural Language Inference: ANLI R1, R2, R3, CB, RTE \cite{anli, RTE}. \newline{} 2. Coreference Resolution: WSC, Winogrande.\cite{WinogradSchema}\newline{} 3. Sentence Completion: COPA\cite{Copa} , HellaSwag \cite{hellaswag}. \newline{} 4. Word Sense Disambiguation: WiC \cite{WiC}. \\
2 & GRIPS \cite{grips} & 8 classification tasks from NaturalInstructions \cite{naturalinstructions} \\
3 & Instruction induction \cite{InstructionInduction} & 1. Spelling, 2. Syntax, 3. Morpho-syntax, 4. Lexical semantics, \newline{}5. Phonetics, 6. Knowledge, 7. Semantics, 8. Style \\
4 & RLPrompt \cite{rlprompt} & 1. Classification \newline{} 2. Text-style transfer \\
5 & TEMPERA \cite{tempera} & Classification \\
6 & AELP \cite{aelp} & Big Bench Hard \cite{bbh} \\
7 & APE \cite{ape} & 1. 24 Instruction induction tasks \cite{InstructionInduction} 2. 21 BIG Bench Hard tasks \cite{bbh} \\
8 & AutoHint \cite{hint} & BIG-Bench Instruction Induction (Epistemic Reasoning, Logical Fallacy Detection, Implicatures, Hyperbaton, Causal Judgment, Winowhy) \cite{ape} \\
9 & BDPL \cite{bdpl} & 1. MNLI \cite{mnli}, 2. QQP \cite{QQP}, 3. SST-2 \cite{sst2}, 4. MRPC \cite{MRPC}, 5. CoLA \cite{cola}, 6. QNLI \cite{qnli}, 7. RTE \cite{RTE}, 8. CitationIntent \cite{citationintent}, 9. SciERC \cite{sciERC}, 10. RCT \cite{RCT}, 11. HyperPartisan \cite{HyperPartisan} \\
10 & Boosted Prompting \cite{boosted-prompting} & GSM8K \cite{gsm8k} and AQuA \cite{AQuA} \\
11 & BPO \cite{bpo-cheng} & Generation: Dolly Eval \cite{dollyEval}, Vicuna Eval \cite{VicunaEval}, Self-Instruct Eval \cite{SelfInstructEval} \\
12 & CLAPS \cite{claps} &  \\
13 & Directional-stimulus \cite{directional-stimulus} & MultiWOZ \cite{multiwoz} \\
14 & DLN \cite{dln} & 1. Mpqa Sentiment analysis \cite{mpqatrecsubj} \newline{} 2. Trec Question type classification \cite{mpqatrecsubj} \newline{}3. Subj Determine whether a sentence is subjective or objective \cite{mpqatrecsubj} \newline{}4. Leopard \cite{leopard}- Disaster Determine whether a sentence is relevant to a disaster. \newline{}5. Leopard \cite{leopard}- Airline Airline tweet sentiment analysis. \newline{}6. BBH \cite{bbh}- (Hyper, Nav, Date, Logic datasets) \\
15 & DSP \cite{dsp} & 1. open-domain question answering (Open-SQuAD) \cite{open-squad} \newline{} 2. multi-hop question answering (HotPotQA) \cite{hotpotqa} \newline{}3. conversational question answering (QReCC) \cite{QReCC} \\
16 & DSPy \cite{dspy} &  \\
17 & GATE \cite{gate} & LAPS \cite{gate} (1. Content Recommendation (user likes to read a given held-out article or not) 2. Moral Reasoning, 3. Email Verification) \\
18 & GPO \cite{gpo} & 1. Sentiment analysis - Yelp \cite{zhang2015character}, Flipkart \cite{flipkart}, IMDB \cite{imdb}, Amazon \cite{zhang2015character} \newline{} 2. NLI - MNLI \cite{mnli}, ANLI \cite{anli} 3.Entailment - RTE \cite{RTE}, 4. CommonsenseQA - SocialIQA \cite{socialIQA} \newline{} 5. Multi-turn dialog - DSTC7 \cite{DSTC7}, Ubuntu Dialog \cite{UbuntuDialog}, MuTual \cite{MuTual} \newline{} 6. NumericalQA - DROP \cite{DROP} \\
19 & PACE \cite{pace} & BBH \cite{bbh}, instruction induction tasks (24 tasks) \cite{InstructionInduction} and translation tasks (en-de, en-es, en-fr) \\
20 & PREFER \cite{prefer} & 1. NLI tasks including SNLI \cite{snli}, MNLI \cite{mnli}, QNLI \cite{qnli}, RTE \cite{RTE} \newline{} 2. Classification: Ethos \cite{ethos}, liar \cite{liar}, ArSarcasm \cite{ArSarcasm} \\
21 & Promptagent \cite{promptagent} & 1. BigBenchHard (BBH) \cite{bbh} - 6 BBH tasks that emphasize a blend of domain knowledge\newline{} 2. Biomedical - Disease NER (NCBI) \cite{ncbi}, MedQA \cite{medQA}, Bio similar sentences \cite{biosimilar} \newline{} 3. 2 classification - TREC \cite{trec} + Subj. \cite{subj} 1 NLI(CB) \cite{nli-ci} \\
22 & Promptboosting \cite{prompt-boosting} & Text Classification \\
23 & Promptbreeder \cite{prompt-breeder} & 1. Arithmetic Reasoning: Benchmarks: GSM8K \cite{gsm8k}, MultiArith \cite{MultiArith}, AddSub \cite{AddSub}, \newline{}SVAMP \cite{Svamp}, SingleEq \cite{SingleEQ}, AQuA-RAT \cite{AquaRat}. \newline{} 2. Commonsense Reasoning: Benchmarks: CommonSenseQA (CSQA) \cite{CSQA}, StrategyQA (SQA) \cite{SQA}. \newline{} 3. Hate Speech Classification: Dataset: ETHOS \cite{ethos}. \newline{} 4. Instruction Induction \cite{InstructionInduction}: Tasks: 24 datasets spanning \newline{} sentence similarity, style transfer, sentiment analysis, and more \\
\hline
\end{tabular}}
\caption{Tasks covered in the different papers}
\end{table*}

\begin{table*}[ht]
\rowcolors{2}{gray!15}{white} 
\resizebox{1\textwidth}{!}{%
\begin{tabular}{|ccp{0.9\textwidth}|}%
\hline
\textbf{SNo.} & \textbf{Paper} & \textbf{Tasks}\\
\hline

24 & ProTeGi \cite{protegi} & Jailbreak \cite{protegi}, Liar \cite{liar}, Sarcasm \cite{Sarcasm}, Ethos \cite{ethos} \\
25 & Random separators \cite{random-separators} & 1. SST-2, SST-5,\cite{sst2} 3. DBPedia \cite{zhang2015character}, 4. MR \cite{MR}, 5. CR \cite{CR}, 6. MPQA \cite{MPQA}, 7. Subj \cite{subj}, 8. TREC \cite{trec}, 9. AGNews \cite{zhang2015character} \\
26 & ABO \cite{abo} & BigBenchHard tasks \cite{bbh}: Object Counting, Navigate, Snarks, Question Selection \\
27 & Adv-ICL \cite{adv-icl} & Summarization (XSUM \cite{XSUM}, CNN/Daily Mail \cite{CNNDailyMail}), Data-to-Text (WebNLG \cite{WebNLG}, E2E NLG \cite{E2ENLG}), Translation (LIRO \cite{LiRO}, TED Talks \cite{TedTalks}), Classification (YELP-5 \cite{zhang2015character}, WSC \cite{WinogradSchema}), Reasoning (GSM8k \cite{gsm8k}, SVAMP \cite{Svamp}) \\
28 & AMPO \cite{ampo} & Text classification task TREC \cite{trec}, \newline{}sentiment classification task SST-5 \cite{sst2}, \newline{}largescale reading comprehension task RACE \cite{RACE}, \newline{} medical question-answering tasks MedQA \cite{medQA} and MedMCQA \cite{MedMCQA} \\
29 & APEER \cite{apeer} & Passage reranking \\
30 & APOHF \cite{apohf} & 1. User instruction optimization using tasks from Instructzero, 2. Text-to-image , 3. Response optimization \\
31 & BATPrompt \cite{bat-prompt} & 1. Language understanding, 2. Text summarization, 3. Text simplification \\
32 & COPLE \cite{cople} & GLUE - SST2 \cite{sst2}, COLA \cite{cola}, MNLI \cite{mnli}, QNLI \cite{qnli}, RTE \cite{RTE}, MRPC \cite{MRPC}, QQP \cite{QQP}  MMLU \cite{MMLU} - STEM, Humanities, Social Sciences and Other \\
33 & CRISPO \cite{crispo} & Summarization, QA \\
34 & DAPO \cite{dapo} & 1. Sentiment classification, 2. topic classification, 3. News, 4. TREC \cite{trec}, 5. subjectivity classification \cite{subj}, 6. Logic Five, 7. Hyperbaton, 8. Disambiguation, 9. Salient, 10.Translation \\
35 & DRPO \cite{drpo} & Alignment benchmark \\
36 & EVOPROMPT \cite{evoprompt} & 1. Language Understanding: Sentiment classification (e.g., SST-2, SST-5, CR, MR \cite{sst2,CR, MR}), 2. Topic classification (e.g., AGNews \cite{zhang2015character}, TREC \cite{trec}), Subjectivity classification (Subj \cite{subj}). 3. Language Generation: Summarization (SAMSum \cite{samsum}). Simplification (ASSET \cite{asset}). 4. Reasoning (BIG-Bench Hard Tasks) \cite{bbh}: Multi-step reasoning tasks from BBH, such as logical deduction, causal judgment, and object tracking. \\
37 & FIPO \cite{fipo} & 1. Generation: GSM8K \cite{gsm8k}, BBH \cite{bbh} 2. Multiple Choice: PiQA \cite{piqa}, CosmosQA \cite{cosmosqa}, MMLU \cite{MMLU} \\
38 & LMEA \cite{lmea} & Traveling Salesman Problems (TSPs) \\
39 & MIPRO \cite{mipro} & 1. Question Answering (HotPotQA)\cite{hotpotqa}  2. Classification (Iris \cite{iris}, Heart Disease \cite{heartdieases}) 3. Entailment (ScoNe) \cite{scone} 4. Multi-hop Fact Extraction and Claim Verification (HoVer) \cite{hover} \\
40 & MOP \cite{mop} & 50 tasks comprising of Instruction Induction \cite{InstructionInduction}, Super Natural Instructions \cite{naturalinstructions}, BBH \cite{bbh} \\
41 & MORL-Prompt \cite{morl-prompt} & 1. Unsupervised Text Style Transfer: Shakespearean data \cite{shakespeare} 2. Supervised Machine Translation: iwslt2017 \cite{iwslt} \\
42 & OIRL \cite{oirl} & Arithmetic reasoning: GSM8K \cite{gsm8k}, MAWPS, SVAMP \cite{Svamp} \\
43 & OPRO \cite{opro} & GSM8K \cite{gsm8k}, BBH (23 tasks) \cite{bbh}, MultiArith \cite{MultiArith}, AQuA \cite{AQuA} \\
44 & PE2 \cite{pe2} & 1. MultiArith and GSM8K for math reasoning \cite{gsm8k}, \newline{} 2. Instruction Induction \cite{InstructionInduction}, \newline{} 3. BIG-bench Hard for challenging LLM tasks \cite{bbh} \newline{} 4. Counterfactual Evaluation \newline{} 5. Production Prompt \\
45 & PIN \cite{pin} & 1. Classification: SST-2 and etc \cite{sst2} \newline{} 2. Unsupervised Text Style transfer: Yelp \cite{zhang2015character}\newline{} 3.Textual Inversion From Images: MSCOCO \cite{mscoco}, LAION \cite{laion} \\
46 & PLUM \cite{plum} & Natural-Instructions datasets v2.6 \cite{naturalinstructions} \\
47 & PRewrite \cite{prewrite} & 1. Classification: AG News \cite{zhang2015character}, SST-2 \cite{sst2}\newline{} 2. Question answering: NQ \cite{NQ}\newline{} 3. Arithmetic reasoning: GSM8K \cite{gsm8k} \\
48 & PROMPTWIZARD \cite{prompt-wizard} & 1. BIG-Bench Instruction Induction (BBII)  \cite{InstructionInduction} \newline{} 2. GSM8k \cite{gsm8k}, AQUARAT \cite{AquaRat}, and SVAMP \cite{Svamp} \newline{} 3. BIG-Bench Hard (BBH) \cite{bbh} \newline{} 4. MMLU \cite{MMLU}, Ethos \cite{ethos}, PubMedQA \cite{pubmedqa}, MedQA \cite{medQA} \\
49 & PROMST \cite{promst} & 11 multistep tasks: 1. Webarena, 2. Alfworld \cite{alfworld}, 3. Scienceworld \cite{scienceworld}, 4. BoxNet1 \cite{boxnet}, 5. BoxNet2, \newline{}6. BoxLift, 7. Warehouse, 8. Gridworld 1, 9. Gridworld 2, 10. Blocksworld, 11. Logistics \\
50 & Reprompting \cite{reprompting} & BBH \cite{bbh}, GSM8K \cite{gsm8k}, MATH \cite{MATH}\\

\hline
\end{tabular}}
\caption{Tasks covered in the different papers}
\end{table*}

\begin{table*}[ht]
\rowcolors{2}{gray!15}{white} 
\resizebox{1\textwidth}{!}{%
\begin{tabular}{|ccp{0.9\textwidth}|}%
\hline
\textbf{SNo.} & \textbf{Paper} & \textbf{Tasks}\\
\hline
51 & SAMMO \cite{sammo} & 1. BigBench zero-shot classification tasks \cite{bigbench} \newline{} 2. GeoQuery \cite{geoquery}, SMCalFlow \cite{smcalflow}, Overnight \cite{overnight} 3. Super-NaturalInstructions \cite{naturalinstructions} \\
52 & SCULPT \cite{sculpt} & BBH (23 tasks) \cite{bbh}, RAI \cite{sculpt} \\
53 & SOS \cite{sos} & 1. Sentiment Analysis  2. Orthography Analysis, 3. Taxonomy of Animals, 4. Disambiguation QA, 5. Logical Five, 6. Color Reasoning \\
54 & SPRIG \cite{sprig} & 1. Reasoning: Tasks requiring multi-step logic or causal reasoning. \newline{} 2. Math: Arithmetic and logical deduction problems. \newline{} 3. Social Understanding: Empathy detection, humor identification, and politeness evaluation. \newline{} 4. Commonsense: Inference tasks like object counting and temporal reasoning. \newline{} 5. Faithfulness: Ensuring generated outputs align with input data. \newline{} 6. Knowledge: Open-domain QA and knowledge recall tasks. \newline{} 7. Language Understanding: Tasks like sentiment analysis and text classification.\newline{} 8. Popular benchmarks include MMLU \cite{MMLU}, BBH \cite{bbh}, TruthfulQA \cite{truthfulqa}, XCOPA \cite{xcopa}, SocKET \cite{socket}, and others, covering 47 task types across multiple languages and domains. \\
55 & StraGo \cite{strago} & BBH \cite{bbh}(five challenging tasks within Big-Bench Hard)  2. SST-5 \cite{sst2}(fine-grained sentiment classification)  3. TREC \cite{trec}(question-type classification). 4. MedQA \cite{medQA},MedMCQA \cite{MedMCQA} (medical-domain QA)  5. Personalized Intent Query (an internal industrial scenario) \\
56 & TextGrad \cite{textgrad} & LeetCode Hard \cite{LeetcodeHard}, Google-proof QA \cite{GoogleProofQA}, MMLU \cite{MMLU} (Machine Learning, College Physics), BBH \cite{bbh} (Object Counting, Word Sorting), GSM8k \cite{gsm8k}, DOCKSTRING \cite{Dockstring}(molecule evaluation)\\
57 & UNIPROMPT \cite{uniprompt} & (1) Ethos \cite{ethos}, (2) ARC \cite{ARC} , (3) MedQA \cite{medQA}, (4) GSM8K \cite{gsm8k} and (5) one real-world task: Search Query Intent \cite{uniprompt} \\

\hline
\end{tabular}}
\caption{Tasks covered in the different papers}
\end{table*}
\clearpage
\section{Prompt examples} 
\subsection{Instruction Induction}
\label{appendix:instruction_induction}
Below is the original instruction induction prompt used by \citet{instruction-induction}
\begin{mdframed}[backgroundcolor=gray!20, linecolor=black]
\{\{\# system $\sim$ \}\} \\
You are a helpful assistant\\
\{\{$\sim/$ system \}\} \\
\{\{\# user $\sim$\}\} \\
I gave a friend an instruction and [[n\_demo]] inputs. The friend read the instruction and wrote an output for every one of the inputs.
Here are the input - output pairs:\\
\{\{ demos \}\} \\
What was the instruction ? It has to be less than \{\{ max\_tokens \}\} tokens . \\
\{\{$\sim/$ user \}\} \\
\{\{\# assistant $\sim$\}\}\\
The instruction was \{\{gen 'instruction ' [[ GENERATION\_CONFIG ]]\}\}\\
\{\{$\sim/$ assistant \}\}
\end{mdframed}

\subsection{Metaprompt design example}
\label{appendix:metaprompt}
Below is the metaprompt used in OPRO \cite{opro}
\begin{mdframed}[backgroundcolor=gray!20, linecolor=black]
I have some texts along with their corresponding scores. The texts are arranged in ascending order based on their scores, where higher scores indicate better quality. text: \\Let’s figure it out! \\score: 61 \\text: Let’s solve the problem. \\score: 63 \\(. . . more instructions and scores . . . ) \\The following exemplars show how to apply your text: \\you replace in each input with your text, then read the input and give an output. We say your output is wrong if your output is different from the given output, and we say your output is correct if they are the same. \\input: Q: Alannah, Beatrix, and Queen are preparing for the new school year and have been given books by their parents. Alannah has 20 more books than Beatrix. Queen has 1/5 times more books than Alannah. If Beatrix has 30 books, how many books do the three have together? \\A: output: 140 \\(. . . more exemplars . . . ) \\Write your new text that is different from the old ones and has a score as high as possible. Write the text in square brackets
\end{mdframed}

\subsection{LLM Feedback prompts} 
\label{appendix:tab_llmaaj}
\begin{sidewaystable}[h]
\centering
\caption{Automatic prompt optimization for LLM-as-a-Judge methods, text gradients~\cite{protegi,promptagent} and PE2~\cite{pe2}.}
\vspace{0pt}
\resizebox{\textwidth}{!}{
\begin{tabular}{lllllll}
\toprule
\begin{tabular}[c]{@{}l@{}l} Method \\ \, \\ \end{tabular}
& \begin{tabular}[c]{@{}l@{}l} LLMaaJ prompt \\ \, \\ \end{tabular}
& \begin{tabular}[c]{@{}l@{}l} Candidate prompt \\ \, \\ \end{tabular}
& \begin{tabular}[c]{@{}l@{}l} Response \\ \, \\ \end{tabular}
& \begin{tabular}[c]{@{}l@{}l} Subject of evaluation \\ (prompt / response / both) \\ \end{tabular}
& \begin{tabular}[c]{@{}l@{}l} Evaluation output \\ \, \\ \end{tabular}
& \begin{tabular}[c]{@{}l@{}l} Rewritten prompt \\ \, \\ \end{tabular}  \\
\midrule 
Text-gradients~\cite{protegi} & 
\begin{tabular}[c]{@{}p{0.2\linewidth}@{}}  
I'm trying to write a zero-shot classifier prompt. My current prompt is:\\
"\{prompt\}" \\
But this prompt gets the following examples wrong:\\
\{error\_string\} \\
give \{num\_feedbacks\} reasons why the prompt could have gotten these examples wrong. Wrap each reason with <START> and <END>
\end{tabular} &
\begin{tabular}[c]{@{}p{0.2\linewidth}@{}}  
Determine whether the Statement is a lie (Yes) or not (No) based on the Context and other information.\\
Statement: Small businesses (are) going out of business in record numbers. Job title: Senator. State: Texas. Party: republican. Context: a speech at Liberty University" \\
Label: Yes Prediction: No
\end{tabular} &
\begin{tabular}[c]{@{}p{0.1\linewidth}@{}}  
N/A
\end{tabular} &
\begin{tabular}[c]{@{}p{0.1\linewidth}@{}}  
Prompt
\end{tabular} &
\begin{tabular}[c]{@{}p{0.2\linewidth}@{}}  
The prompt does not take into account the speaker’s potential biases or agenda, which could influence the veracity
of their statements.
\end{tabular} &
\begin{tabular}[c]{@{}p{0.2\linewidth}@{}}  
Determine if the statement is true (Yes) or false (No) based on the context, sources referenced, and potential
biases of the speaker.
\end{tabular}
\\
\midrule

Text-gradients~\cite{promptagent} & 
\begin{tabular}[c]{@{}p{0.2\linewidth}@{}}  
I’m writing prompts for a language model designed for a task. My current prompt is: \\
\{cur prompt\} \\
But this prompt gets the following examples wrong: \\
\{error string\} \\
For each wrong example, carefully examine each question and wrong, answer step by step, provide comprehensive and different reasons why
the prompt leads to the wrong answer. At last, based on all these reasons, summarize and list all the aspects that can improve the prompt.
\end{tabular} &
\begin{tabular}[c]{@{}p{0.2\linewidth}@{}}  
Premise: William learns that kids play in water coming up in streams out of a tiled floor with image of a large rose on it.\\
Hypothesis: William learns that kids are playing in water.\\
Label: Non-entailment Prediction: Entailment
\end{tabular} &
\begin{tabular}[c]{@{}p{0.1\linewidth}@{}}  
Non-entailment
\end{tabular} &
\begin{tabular}[c]{@{}p{0.1\linewidth}@{}}  
Prompt
\end{tabular} &
\begin{tabular}[c]{@{}p{0.2\linewidth}@{}}  
Error Feedback: "Ignoring context and detail" The model might be overlooking the details of the premise 'kids play in water coming up in
streams out of a tiled floor with an image of a large rose on it,', which directly implies the hypothesis.
\end{tabular} &
\begin{tabular}[c]{@{}p{0.2\linewidth}@{}}  
Compare the provided sentences. Take into account the subtleties in the context, pinpoint the order of events and differentiate between facts and assumptions. If the hypothesis is a direct result of the premise, select 'entailment'.
\end{tabular}
\\

\midrule
PE2~\cite{pe2} & 
\begin{tabular}[c]{@{}p{0.2\linewidth}@{}}  
\# Instruction For each example, provide reasoning according to the following template \\
* Output is correct? \\
* Necessary to edit the prompt? \\
* If yes, suggestions on prompt editing?
\end{tabular} &
\begin{tabular}[c]{@{}p{0.2\linewidth}@{}}  
\# Current Prompt Let’s think step by step. \# Full Template ``` Question: Answer: Let’s think step by step. ``` \# Examples \#\# Example 1 Input: George had 28 socks. If he threw away 4 socks ... Output: 64 Reasoning: Step 1: George had 28 socks. Step 2: ... Label: 60 [More examples ...]
\end{tabular} &
\begin{tabular}[c]{@{}p{0.1\linewidth}@{}}  
N/A
\end{tabular} &
\begin{tabular}[c]{@{}p{0.1\linewidth}@{}}  
Both
\end{tabular} &
\begin{tabular}[c]{@{}p{0.2\linewidth}@{}}  
\#\# Example 1 Output is correct? No. Reasoning: the model didn't subtract the socks he threw away. Prompt describing the task correctly? Yes. Necessary to edit the prompt? Yes. Suggestions: The prompt should be edited to guide the model to perform subtraction. [More examples ...]
\end{tabular} &
\begin{tabular}[c]{@{}p{0.2\linewidth}@{}}  
Now carefully review your reasoning and proceed with step 2: refine the prompt. \# Current Prompt Let’s think step by step. \# Instructions * The total length should be less than 50 words * Reply with the prompt. Do not include other text.
\end{tabular}
\\

\bottomrule
\end{tabular}
}
\label{tab:llmaaj_part1}
\vspace{0pt}
\end{sidewaystable}

\begin{sidewaystable}[h]
\centering
\caption{Automatic prompt optimization for LLM-as-a-Judge methods, Hints~\cite{hint}.}
\vspace{0pt}
\resizebox{\textwidth}{!}{
\begin{tabular}{lllllll}
\toprule
\begin{tabular}[c]{@{}l@{}l} Method \\ \, \\ \end{tabular}
& \begin{tabular}[c]{@{}l@{}l} LLMaaJ prompt \\ \, \\ \end{tabular}
& \begin{tabular}[c]{@{}l@{}l} Candidate prompt \\ \, \\ \end{tabular}
& \begin{tabular}[c]{@{}l@{}l} Response \\ \, \\ \end{tabular}
& \begin{tabular}[c]{@{}l@{}l} Subject of evaluation \\ (prompt / response / both) \\ \end{tabular}
& \begin{tabular}[c]{@{}l@{}l} Evaluation output \\ \, \\ \end{tabular}
& \begin{tabular}[c]{@{}l@{}l} Rewritten prompt \\ \, \\ \end{tabular}  \\

\midrule

Hints~\cite{hint} & 
\begin{tabular}[c]{@{}p{0.2\linewidth}@{}}  
Given following task: [Task Description] \\
Given Input: [Input] \\
And its expected Output: [output] \\
List the reason or hint why it's with this expected output within tag <hint> and </hint>.
\end{tabular} &
\begin{tabular}[c]{@{}p{0.2\linewidth}@{}}  
Determine whether one sentence entails the next \\
\# Given Input: [input] \\
Identify the relation between the following premises and hypotheses, choosing from the options 'entailment' or 'non-entailment'. \\
Put your answer within tag <Ans> and </Ans>. \\
\# Result

\end{tabular} &
\begin{tabular}[c]{@{}p{0.1\linewidth}@{}}  
Non-entailment
\end{tabular} &
\begin{tabular}[c]{@{}p{0.1\linewidth}@{}}  
Prompt
\end{tabular} &
\begin{tabular}[c]{@{}p{0.2\linewidth}@{}}  
- Entailment occurs when the hypothesis is a logical consequence of the premise, or when the premise guarantees the truth of the hypothesis, regardless of the level of specificity or simplification of the terms involved. \\
- Non-entailment occurs when the premise does not guarantee the truth of the hypothesis, or when there is a possibility that the hypothesis is false or unknown, especially when the premise involves beliefs or thoughts of other people. 
\end{tabular} &
\begin{tabular}[c]{@{}p{0.2\linewidth}@{}}  
Determine whether one sentence entails the next. Some useful hints are:\\
- Entailment occurs when the hypothesis is a logical consequence of the premise, or when the premise guarantees the truth of the hypothesis, regardless of the level of specificity or simplification of the terms involved.\\
- Non-entailment occurs when the premise does not guarantee the truth of the hypothesis, or when there is a possibility that the hypothesis is false or unknown, especially when the premise involves beliefs or thoughts of other people.\\
\# Given Input: [input]\\
Identify the relation between the following premises and hypotheses, choosing from the options 'entailment' or 'non-entailment'. Put your answer within tag <Ans> and </Ans>.\\
\# Result

\end{tabular}
\\

\bottomrule
\end{tabular}
}
\label{tab:llmaaj_part2}
\vspace{0pt}
\end{sidewaystable}

\begin{sidewaystable}[h]
\centering
\caption{Automatic prompt optimization for LLM-as-a-Judge methods, Critique~\cite{crispo}.}
\vspace{0pt}
\resizebox{\textwidth}{!}{
\begin{tabular}{lllllll}
\toprule
\begin{tabular}[c]{@{}l@{}l} Method \\ \, \\ \end{tabular}
& \begin{tabular}[c]{@{}l@{}l} LLMaaJ prompt \\ \, \\ \end{tabular}
& \begin{tabular}[c]{@{}l@{}l} Candidate prompt \\ \, \\ \end{tabular}
& \begin{tabular}[c]{@{}l@{}l} Response \\ \, \\ \end{tabular}
& \begin{tabular}[c]{@{}l@{}l} Subject of evaluation \\ (prompt / response / both) \\ \end{tabular}
& \begin{tabular}[c]{@{}l@{}l} Evaluation output \\ \, \\ \end{tabular}
& \begin{tabular}[c]{@{}l@{}l} Rewritten prompt \\ \, \\ \end{tabular}  \\

\midrule

Critique~\cite{crispo} & 
\begin{tabular}[c]{@{}p{0.2\linewidth}@{}}  
Critique:\\
- Number of words: The predicted summaries tended to be longer with more details while the reference summaries were shorter and more concise.\\
- Number of sentences: The predicted summaries used more sentences to describe the inputs while the reference summaries were more succinct with fewer sentences.\\
- Precision: Some details in the predicted summaries were not important and not mentioned in the reference summaries\\
- Recall: Some key details highlighted in the reference summaries were missing from the predicted summaries.\\

Suggestion:\\
- Specifying the expected length of the summary (e.g. 1-2 sentences)\\
- Emphasizing to only include the most important/key details\\
- Indicating which details should be prioritized or omitted

\end{tabular} &
\begin{tabular}[c]{@{}p{0.2\linewidth}@{}}  
For the given text, write a 1-2 sentence summary within 〈summary〉 tags that highlights the most important details. Focus on including who the key people are and what happened between them.\\
INSERT INPUT HERE\\
Some key details to focus on include the main characters, any plans or arrangements that were made, and the overall outcome or resolution.
\end{tabular} &
\begin{tabular}[c]{@{}p{0.1\linewidth}@{}}  
Tegan tells Valentia that Paul’s brother sent her a friend request on social media, though she and Paul had previously broken up. 
Valentia advises Tegan to ignore the request, not wanting Tegan to revisit her past relationship with Paul.  \\
Score: 42.1
\end{tabular} &
\begin{tabular}[c]{@{}p{0.1\linewidth}@{}}  
both
\end{tabular} &
\begin{tabular}[c]{@{}p{0.2\linewidth}@{}}  
Comparing the high-score and low-score instructions, here are some suggestions that could improve them:\\
〈suggestion〉 Specify the desired length or range for the summaries (e.g., 10 words and 1-2 sentences).〈/suggestions〉\\
〈suggestion〉 Specify to focus on key events and specify which details 〈/suggestion〉\\
〈suggestion〉 Specify the output should not contain unnessary context 〈/suggestion〉
\end{tabular} &
\begin{tabular}[c]{@{}p{0.2\linewidth}@{}}  
Read the dialogue provided in INSERT INPUT HERE and identify the key events between characters and outcomes. Then write a 1-2 sentence
summary within 〈summary〉 tags that concisely captures these important plot points, such as who will borrow a dress or who has an interview, while keeping within 10 words where possible. Focus only on the characters and salient events, omitting unnecessary context.
\end{tabular}
\\

\bottomrule
\end{tabular}
}
\label{tab:llmaaj_part3}
\vspace{0pt}
\end{sidewaystable}

\begin{sidewaystable}[h]
\centering
\caption{Automatic prompt optimization for LLM-as-a-Judge methods, Reflection~\cite{cieri-etal-2022-reflections}.}
\vspace{0pt}
\resizebox{\textwidth}{!}{
\begin{tabular}{lllllll}
\toprule
\begin{tabular}[c]{@{}l@{}l} Method \\ \, \\ \end{tabular}
& \begin{tabular}[c]{@{}l@{}l} LLMaaJ prompt \\ \, \\ \end{tabular}
& \begin{tabular}[c]{@{}l@{}l} Candidate prompt \\ \, \\ \end{tabular}
& \begin{tabular}[c]{@{}l@{}l} Response \\ \, \\ \end{tabular}
& \begin{tabular}[c]{@{}l@{}l} Subject of evaluation \\ (prompt / response / both) \\ \end{tabular}
& \begin{tabular}[c]{@{}l@{}l} Evaluation output \\ \, \\ \end{tabular}
& \begin{tabular}[c]{@{}l@{}l} Rewritten prompt \\ \, \\ \end{tabular}  \\

\midrule

Reflection~\cite{cieri-etal-2022-reflections} & 
\begin{tabular}[c]{@{}p{0.2\linewidth}@{}}  
Here is a conversation with an LLM: \\
\{x|y\}. \\
Below are the criticisms on \{y\}: \\
Explain how to improve \{x\}.

\end{tabular} &
\begin{tabular}[c]{@{}p{0.2\linewidth}@{}}  
Below are the criticisms on \{x\}:  \\
Incorporate the criticisms, and produce a new variable.
\end{tabular} &
\begin{tabular}[c]{@{}p{0.1\linewidth}@{}}  
N/A
\end{tabular} &
\begin{tabular}[c]{@{}p{0.1\linewidth}@{}}  
both
\end{tabular} &
\begin{tabular}[c]{@{}p{0.2\linewidth}@{}}  
Exmaple output for instance optimization (a specific coding problem, for example):\\
Handling `nums[i] == k`**: The current logic does not correctly handle the case when `nums[i] == k`. The balance should be reset or adjusted differently when `k` is encountered. \\
Output for prompt optimization:\\
The evaluator LLM simply returns if the generated output and ground truth matches (math problem result, for example)
\end{tabular} &
\begin{tabular}[c]{@{}p{0.2\linewidth}@{}}  
For prompt optimization:\\
From: You will answer a reasoning question. Think step by step. The last line of your response should be of the following format: 'Answer: \$VALUE' where VALUE is a numerical value.\\

To: You will answer a reasoning question. List each item and its quantity in a clear and consistent format, such as '- Item: Quantity'. Sum the values directly from the list and provide a concise summation. Ensure the final answer is clearly indicated in the format: 'Answer: \$VALUE' where VALUE is a numerical value. Verify the relevance of each item to the context of the query and handle potential errors or ambiguities in the input. Double-check the final count to ensure accuracy."
\end{tabular}
\\

\bottomrule
\end{tabular}
}
\label{tab:llmaaj_part4}
\vspace{0pt}
\end{sidewaystable}

\begin{sidewaystable}[h]
\centering
\caption{Automatic prompt optimization for LLM-as-a-Judge methods, Safety Score~\cite{sos}.}
\vspace{0pt}
\resizebox{\textwidth}{!}{
\begin{tabular}{lllllll}
\toprule
\begin{tabular}[c]{@{}l@{}l} Method \\ \, \\ \end{tabular}
& \begin{tabular}[c]{@{}l@{}l} LLMaaJ prompt \\ \, \\ \end{tabular}
& \begin{tabular}[c]{@{}l@{}l} Candidate prompt \\ \, \\ \end{tabular}
& \begin{tabular}[c]{@{}l@{}l} Response \\ \, \\ \end{tabular}
& \begin{tabular}[c]{@{}l@{}l} Subject of evaluation \\ (prompt / response / both) \\ \end{tabular}
& \begin{tabular}[c]{@{}l@{}l} Evaluation output \\ \, \\ \end{tabular}
& \begin{tabular}[c]{@{}l@{}l} Rewritten prompt \\ \, \\ \end{tabular}  \\

\midrule

Safety Score~\cite{sos} & 
\begin{tabular}[c]{@{}p{0.2\linewidth}@{}}  
You are a security gatekeeper. Given an existing prompt and a series of cases where it made mistakes. Look through each case carefully and identify what is causing the mistakes. Based on these observations, output a way to improve the prompts based on the mistakes. From a security perspective, a mistake was made when the LLM helped the user perform a malicious task. If the LLM responds with a "canned response", like, "I am just an AI, I cannot help you with this..." then the LLM did not help the user. If the LLM says it cannot answer without more information then the LLM did not help.\\
Some ways to make a prompt secure are:\\
1. Add a detailed description of the task and how to solve it. Consider providing a step-by-step description of the process the LLM should follow.\\
2. Specify instructions to ensure the LLM’s answer is respectful.\\
3. Specify guardrails around input and output format.\\
*\{passed in data\}*\\
ways to make the existing prompt more secure based on observations of the mistakes in the cases above are:

\end{tabular} &
\begin{tabular}[c]{@{}p{0.2\linewidth}@{}}  
You are a mutator. Given a prompt, your task is to generate another prompt with the same semantic meaning and intentions.
\# Example:
current prompt: Your mission is to ascertain the sentiment of the provided text and assign a sentiment label from [’negative’, ’positive’].\\
mutated prompt: Determine the sentiment of the given sentence and assign a label from [’negative’, ’positive’].\\
\# Given:\\
\{passed in data\}
\end{tabular} &
\begin{tabular}[c]{@{}p{0.1\linewidth}@{}}  
KPI score: \\
Security score: 
\end{tabular} &
\begin{tabular}[c]{@{}p{0.1\linewidth}@{}}  
Response
\end{tabular} &
\begin{tabular}[c]{@{}p{0.2\linewidth}@{}}  
Select the locally optimal with respect to an objective if it achieves the best performance on this objective among all prompts that exhibit similar performance across all other objectives in other objectives
\end{tabular} &
\begin{tabular}[c]{@{}p{0.2\linewidth}@{}}  
Task Description: You are a Sentiment Classifier. Your task is to classify the emotional sentiment of the text to understand the overall tone and sentiment expressed. This sentiment classification can be used to analyze customer feedback, improve product experiences, or identify potential issues.\\

Instructions for Respectful Language Model: It is important to ensure that the language model promotes positive and respectful interactions. Therefore, please note that the language model should not provide any support or guidance for illegal activities, harmful actions, or unethical behavior. Emphasize the importance of promoting positive and respectful interactions.\\

Input and Output Format: The input should only consist of a single sentence or a short paragraph that requires sentiment classification. The output should provide the emotional sentiment without any additional information or suggestions.\\

Examples:\\
Please classify the emotional sentiment expressed in the following texts:\\
Input: "The new software update is user-friendly and efficient."\\
Correct Answer: [’positive’]\\
\{Another four few shot examples\} \\
Please provide the emotional sentiment for each input text without any additional information or suggestions.
\end{tabular}
\\

\bottomrule
\end{tabular}
}
\label{tab:llmaaj_part5}
\vspace{0pt}
\end{sidewaystable}

\end{document}